\documentclass[12pt]{article}

\usepackage{scicite}

\usepackage{times}

\usepackage{graphicx}
\usepackage{caption}
\graphicspath{{figures/}}
\usepackage{xcolor}
\usepackage{float}
\usepackage{amsmath}
\usepackage{multirow}
\usepackage{pdflscape}
\usepackage{titling}
\usepackage{hyperref}

\newcommand{\commentout}[1]{}

\usepackage{lineno}

\newcommand{\mcot}{c_\text{mt}}

\title{Mechanical Intelligence Simplifies Control in Terrestrial Limbless Locomotion}

\author
{Tianyu Wang,$^{1,2,3,\dagger}$ Christopher Pierce,$^{2,4,\dagger}$ Velin Kojouharov,$^{3}$ Baxi Chong,$^{2}$ \\Kelimar Diaz,$^{2}$ Hang Lu,$^{4}$ Daniel I. Goldman$^{1,2,\ast}$\\
\\
\normalsize{$^{1}$Institute for Robotics and Intelligent Machines, Georgia Institute of Technology,}\\
\normalsize{801 Atlantic Dr NW, Atlanta, GA 30332, USA}\\
\normalsize{$^{2}$School of Physics, Georgia Institute of Technology,}\\
\normalsize{837 State St NW, Atlanta, GA 30332, USA}\\
\normalsize{$^{3}$George W. Woodruff School of Mechanical Engineering, Georgia Institute of Technology,}\\
\normalsize{801 Ferst Dr NW, Atlanta, GA 30318, USA}\\
\normalsize{$^{4}$School of Chemical \&\ Biomolecular Engineering, Georgia Institute of Technology,}\\
\normalsize{311 Ferst Dr, Atlanta, GA 30332, USA}\\
\normalsize{$^\ast$To whom correspondence should be addressed; E-mail: daniel.goldman@physics.gatech.edu.}\\
\normalsize{$^\dagger$These authors contributed equally to this work.}\\ \\
\normalsize{{\color{blue} Published version in Science Robotics: \href{https://doi.org/10.1126/scirobotics.adi2243}{doi.org/10.1126/scirobotics.adi2243}}}
}

\date{}

\begin{document} 


\baselineskip24pt


\maketitle 

\setcounter{figure}{0}
\renewcommand{\figurename}{Fig.}

\section*{Abstract:}
Limbless locomotors, from microscopic worms to macroscopic snakes, traverse complex, heterogeneous natural environments typically using undulatory body wave propagation. Theoretical and robophysical models typically emphasize body kinematics and active neural/electronic control. However, we contend that because such approaches often neglect the role of passive, mechanically controlled processes (those involving ``mechanical intelligence"), they fail to reproduce the performance of even the simplest organisms. To discover principles of how mechanical intelligence aids limbless locomotion in heterogeneous terradynamic regimes, here we conduct a comparative study of locomotion in a model of heterogeneous terrain (lattices of rigid posts). We use a model biological system, the highly studied nematode worm \textit{C. elegans}, and a robophysical device whose bilateral actuator morphology models that of limbless organisms across scales. The robot's kinematics quantitatively reproduce the performance of the nematodes with purely open-loop control; mechanical intelligence simplifies control of obstacle navigation and exploitation by reducing the need for active sensing and feedback. An active behavior observed in \textit{C. elegans}, undulatory wave reversal upon head collisions, robustifies locomotion via exploitation of the systems' mechanical intelligence. Our study provides insights into how neurally simple limbless organisms like nematodes can leverage mechanical intelligence via appropriately tuned bilateral actuation to locomote in complex environments. These principles likely apply to neurally more sophisticated organisms and also provide a design and control paradigm for limbless robots for applications like search and rescue and planetary exploration. 

\section*{Summary:} 
A comparative biological and robotic study reveals principles of mechanical intelligence in terrestrial limbless locomotion. 

\section*{Introduction}
Organisms from flapping hawkmoths~\cite{gau2021rapid} to prancing gazelles~\cite{alexander2003principles} to undulating snakes~\cite{schiebel2019mechanical} and nematodes~\cite{fang2010biomechanical} produce directed movement through a combination of neural and mechanical control. Neural circuits integrate and process sensory information to produce locomotor commands through complex signaling networks. This helps organisms produce directed movement despite the constantly changing external environment by modulating motor commands in response to environmental cues. Much progress has been made in understanding the neural aspects of locomotor control including the structure, function, and dynamics of neural circuits, particularly with genetic models such as \textit{Caenorhabditis elegans}~\cite{white1986structure}, \textit{Drosophila melanogaster}~\cite{chiang2011three}, zebrafish~\cite{naumann2016whole} and mice~\cite{oh2014mesoscale}.

In addition to purely neural control, ``neuromechanical" approaches have been developed to describe the interaction between active neuronal controls and purely mechanical processes arising from body-environment interactions. This approach has been applied primarily to flying and walking systems~\cite{gau2021rapid,sponberg2008neuromechanical,winter2009biomechanics}. For example, fruit flies have been found to recover from flight disturbances through reflexive turning responses to mechanical stimuli~\cite{ristroph2010discovering}, whereas running guineafowl have been shown to stabilize their gaits in rough terrain through passive adaptive responses (``preflexes") mediated by the non-linear properties of the musculature~\cite{daley2006running}. In general, body-environment interactions can help coordinate the movements of the body, through purely mechanical control processes, a phenomenon known as mechanical or physical intelligence~\cite{sitti2021physical}. A complete description of organismal locomotion must therefore place principles of neural/computational intelligence and mechanical intelligence on an equal footing, leading to the concept of embodied intelligence~\cite{pfeifer2007self,pfeifer2006body,iida2022timescales}. 

Although much attention has been paid to mechanical intelligence in legged and aerial systems, less is known about the interplay of neural and mechanical control in limbless locomotion. This locomotor strategy occurs within diverse and often highly complex, heterogeneous environments and spans length scales, from meter-long snakes~\cite{gaymer1971new,jayne1986kinematics,guo2008limbless} with over $10^6$ neurons to the millimeter-long nematode worm \textit{C. elegans} (Fig. 1A), which navigates complex micro-environments like rotting fruit (Fig. 1C, Movie S1) with only 302 neurons~\cite{felix2010natural,cohen2010swimming}. Across the taxonomic and neuroanatomical diversity of lateral undulators, many organisms, including snakes and nematodes, employ similar actuation mechanisms -- bilaterally arranged bands of muscle that propagate waves of contralateral activation down the body, producing undulatory waves that lie in a plane. The ubiquity and biological diversity of undulation, the continual environment-body hydro- and terradynamic interactions, and the existence of common mechanisms of actuation across organisms suggest an important role for mechanical intelligence in limbless locomotion. 

Given the importance of mechanics and the challenges of modeling locomotor-environment interactions, using robots as ``robophysical" models to identify key neuromechanical principles is appealing~\cite{gravish2018robotics,ijspeert2014biorobotics,aguilar2016review}. These models incorporate simplified descriptions of organismal mechanics and neural control, and thus can be used to elucidate the emergent ``template-level" dynamics of organisms~\cite{full1999templates}. This approach has been particularly successful in identifying the role of mechanical control in legged locomotion, including hopping~\cite{meyer2006passive}, bipedal~\cite{collins2005efficient}, quadrupedal~\cite{bledt2018cheetah} and hexapodal~\cite{altendorfer2001rhex} locomotion; and later flapping flight~\cite{dickinson1999wing}. These ``terradynamic" systems have forced researchers to confront the unpredictability, non-linearity and heterogeneity of the physical world. However, these concepts have been less extensively applied in modelling the complex terradynamic interactions and biomechanics of limbless locomotion.

Limbless robots, despite often being referred to as snake-like~\cite{hirose2004biologically,transeth2008snake,crespi2008online,wright2007design}, have yet to match the locomotion capabilities of even the simplest limbless organisms like nematodes. Existing limbless robots which often rely on complex and high bandwidth sensing and feedback~\cite{liu2021review,wang2020directional,sanfilippo2017perception} are stymied by unpredictable terrain in the real world that would not challenge their organismal counterparts~\cite{schiebel2019mechanical,kano2012local,felix2010natural}. Beyond rigid systems, soft limbless robots with intrinsically compliant bodies have emerged over the last decade~\cite{luo2014theoretical,branyan2017soft,qi2022bioinspired}. However, control challenges that arise from air/fluid handling mechanisms and difficulties of modeling and modulating intrinsic material properties have limited their practical uses. Hence limbless robots have yet to fulfill their promised potential for agile movement in the type of complex environments encountered in applications such as search and rescue and planetary exploration. 

One feature of elongated vertebrate and invertebrate organisms that is absent in the direct spinal actuation (joint actuation) design paradigm of limbless robots~\cite{yim2002modular,hirose2004biologically,transeth2008snake,crespi2008online,wright2007design,wu2010cpg,fu2020robotic,takemori2022adaptive} is bilateral actuation. Although simpler in design and control, the joint actuation mechanism limits the usefulness of limbless robotic models in identifying  possible functional roles of bilateral actuation in mechanical control. Indeed, recent work has implied the importance of bilateral actuation in snakes~\cite{schiebel2019mechanical} and limbless devices~\cite{schiebel2020robophysical,boyle2012adaptive,racioppo2019design} when interacting with heterogeneities, suggesting that such an actuation scheme provides a degree of mechanical intelligence and thereby simplifies active control.

To advance our overall understanding and discover principles of mechanical intelligence in limbless locomotion (and to understand the potential role of bilateral actuation specifically in mechanical control), we take a comparative biological and robophysical approach, using two complementary models: a biological model, the nematode \textit{Caenorhabditis elegans}, and a robophysical model, a limbless robot incorporating a bilateral actuation scheme that permits programmable, dynamic, and quantifiable body compliance. This compliance governs the passive body-environment interactions in the horizontal plane that allow mechanical intelligence. Since separating neural and mechanical aspects of control is challenging in a freely locomoting living system, we use the robot as a model~\cite{webb2001can,aguilar2016review,gravish2018robotics,aydin2019physics} which then allows mechanical intelligence to be isolated from active controls and to be systematically tuned and tested. 

\begin{figure}[H]
\centering
\includegraphics[width=0.8\textwidth]{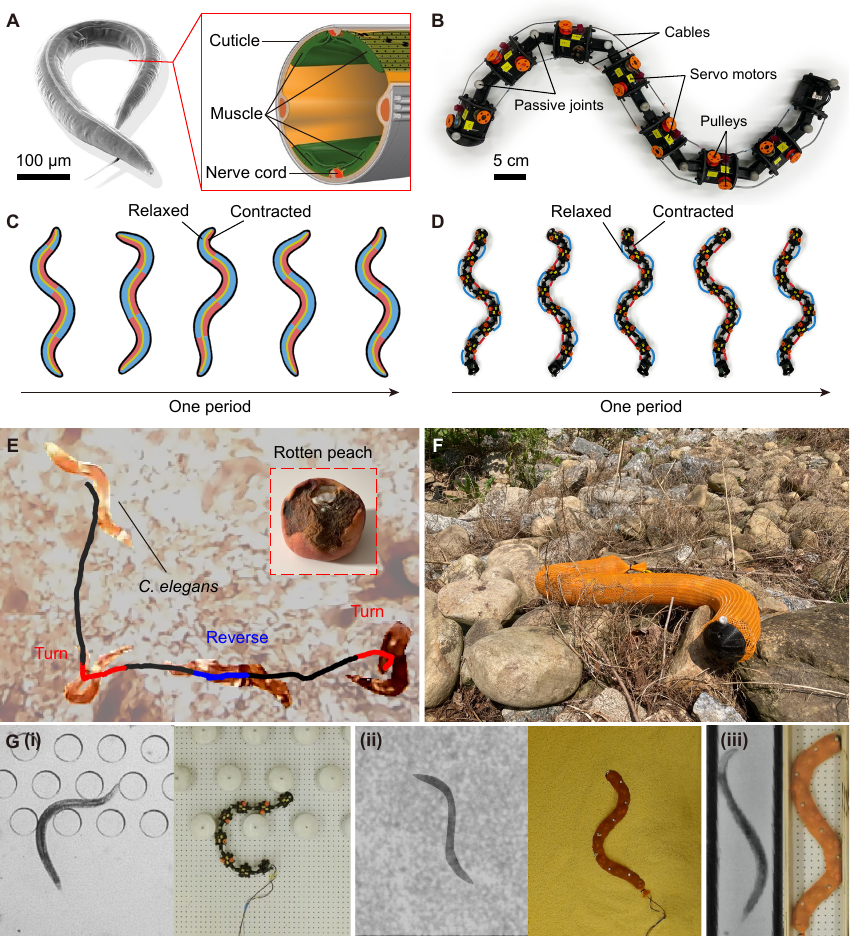}
\caption{Biological and robophysical limbless systems for understanding mechanical intelligence. (\textbf{A}) Nematode \textit{Caenorhabditis elegans}, the biological model of this study (image credit: Ralf J. Sommer), along with a cross-sectional anatomy (reproduced from \cite{WormAtlas}) showing two pairs of bilaterally activated muscle bands. (\textbf{B}) The limbless robophysical model, implementing a bilaterally actuation mechanism. (\textbf{C}) Schematics of body postures and muscle activities over one gait period in the biological model. (\textbf{D}) Schematics of body postures and cable activities over one gait period in the robophysical model. (\textbf{E}) A nematode moves on a slice of the rotten peach, a rheologically complex natural environment. (\textbf{F}) The robophysical model locomotes on a pile of rocks, a rheologically complex natural environment. (\textbf{G}) Biological and robophysical locomotion in comparable laboratory terrestrial environments: (i) lattices, (ii) granular media, and (iii) narrow channels.}
\end{figure}

Using comparisons between the kinematics and locomotor performance of our biological and robophysical models, we will show  that mechanical intelligence alone is sufficient for an open-loop limbless robot to reproduce locomotory behavior of nematodes. Mechanical intelligence simplifies controls in terrestrial limbless locomotion by taking advantage of passive body-environment interactions that enable heterogeneity negotiation, thereby stabilizing locomotion. Further, we show that a simple active behavior inspired by nematodes takes advantage of mechanical intelligence to enhance locomotion performance even further. Our method and results not only provide insight into the functional mechanism of mechanical intelligence in organismal limbless locomotion but also provide an alternative paradigm for limbless robot development that simplifies control in complex environments.

\section*{Results}

\subsection*{Nematode kinematics and performance in heterogeneous terrains}
\textit{C. elegans} (Fig. 1A) has a fully mapped nervous system~\cite{white1986structure,varshney2011structural} with a variety of available genetic tools for perturbing~\cite{fang2015illuminating} and observing~\cite{nguyen2016whole} neuromuscular dynamics. Compared to vertebrate undulators like snakes, its neural control architectures are simpler, and better understood. Moreover, the limited information we have about its ecology and environment suggests it is capable of contending with extremely varied and complex terrain like the interior of rotten fruit~\cite{felix2010natural} (Fig. 1B). Hence it is a promising model for understanding how neural feedback control and mechanical intelligence interact to generate limbless locomotion. We studied \textit{C. elegans} locomotion kinematics using two-dimensional microfluidic hexagonal pillar arrays (or lattices, where pillars are rigid, thus cannot move or deform upon collision with \textit{C. elegans}) with varying pillar density as model heterogeneous environments (Fig. 2A-i, v, ix, Movie S1)~\cite{park2008enhanced,majmudar2012experiments}. These arrays capture aspects of the confinement and potential hindrance to locomotion that natural heterogeneity can impose. Surprisingly, previous work has shown that rather than hindering locomotion, lattices can instead enhance nematode locomotor speeds~\cite{park2008enhanced,majmudar2012experiments}. Moreover, a previous numerical model of a nematode swimming in lattice implicated a strong role for passive mechanics in reproducing the behavior~\cite{majmudar2012experiments}, suggesting that mechanical intelligence likely plays a role in nematodes' ability to take advantage of environmental interactions. However, the detailed kinematics of lattice traversal, particularly during inhibitory head collisions, have not been fully described.

To simplify the analysis of locomotion kinematics in lattices, we exploited dimensionality reduction techniques. Prior work applied principal component analysis to study undulating systems, such as nematodes and snakes, and illustrated that the majority of body postures can be described by linear combinations of sine-like shape-basis functions, despite the inherently high dimensionality of postural data~\cite{rieser2019geometric,chong2022coordinating}. By considering the first two dominant principle components (Fig. 2A-iii, vii, xi), we assumed that the body curvature profile $\kappa$ at time $t$ and location $s$ ($s=0$ denotes head and $s=1$ denotes tail) can be approximated by:
\begin{equation}
\begin{aligned}
\kappa(s,t) &= w_1(t)\sin(2\pi\xi s+\phi) + w_2(t)\cos(2\pi\xi s+\phi)\\
            &= w_1(t)\beta_1(s) + w_2(t)\beta_2(s),
\label{eq:shapeSpaceMap}
\end{aligned}
\end{equation}
where $\xi$ is the spatial frequency of body undulation obtained from direct fitting and $\phi$ is the emergent phase. $w_1(t)$ and $w_2(t)$ are the reduced shape variables describing the instantaneous shape of the locomotor at time $t$. Thus, by projecting curvatures onto the shape-basis functions $\beta_{1,2}(s)$ (Fig. 2A-iii, vii, xi), the locomotion may be visualized as a path (the trajectory formed by $w_1(t)$ and $w_2(t)$) through a two-dimensional ``shape space" defined by $w_1$ and $w_2$ (Fig. 2A-iv, viii, xii, details are provided in Materials and Methods).

We studied nematode locomotion in four environments with varying pillar density, $L/d = 0$ (open fluid), $1.8$ (sparse lattice), $2.8$ (medium lattice), and $3.3$ (dense lattice), where $L$ represents nematode body length and $d$ denotes pillar spacing. Consistent with previous observations~\cite{stephens2008dimensionality}, the nematodes performed an approximate traveling wave motion in homogeneous open fluid. In the shape space, this leads to circular orbits, where one full rotation corresponds to a single undulation cycle (Fig. 2A-i to iv). The nematode maintained a traveling-wave-like gait in all lattice spacings, despite pitch differences. In sparser lattices (Fig. 2A-v to viii), the body kinematics were similar to those in a bulk fluid. Only in the dense lattice (Fig. 2A-ix to xii) did we observe deviations from an ideal travelling wave. However, these deviations were small and transient, so that the overall path in shape space remained mostly circular. These deformations are typically correlated with body deformations induced by collisions (typically between the head and an obstacle) and rapidly ($\sim$0.4 s) return to smooth traveling wave motion. Thus, environmental heterogeneities were observed to induce small perturbations that returned to a stable circular orbit, suggesting that the basic strategy of propagating traveling waves along the body is robust to intrusions by obstacles. 

\begin{figure}[H]
\centering
\includegraphics[width=1\textwidth]{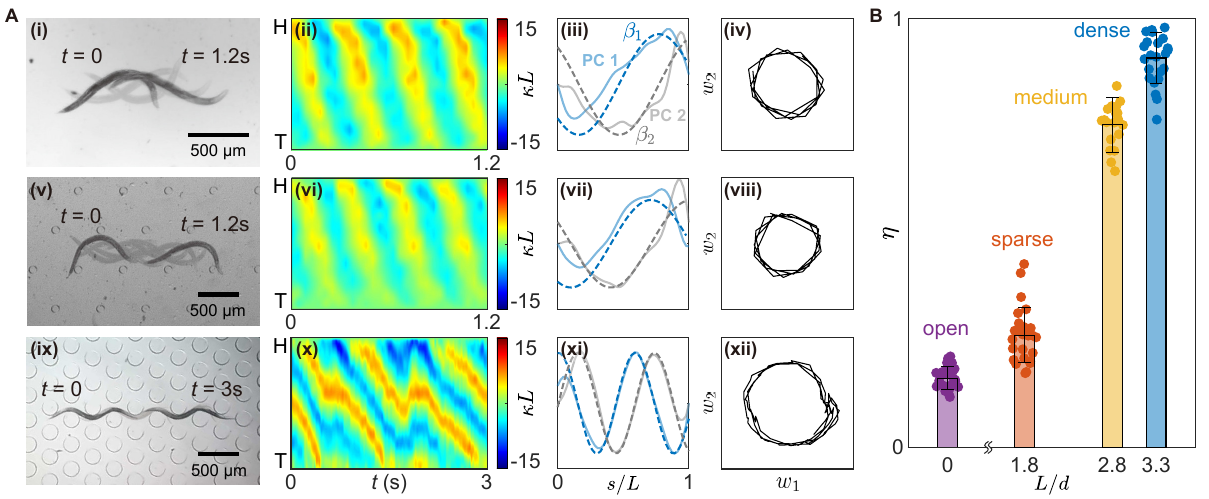}
\caption{Nematode kinematics and performance imply the role of mechanical intelligence. (\textbf{A}) Overlaid snapshots, effective body curvature, gait paths in the shape space, the first two dominant modes (solid lines are the principal components and dashed lines are the best fits to $\sin$ and $\cos$ shape bases) of nematode locomotion in laboratory environments with varied pillar density. (\textbf{B}) Locomotion speed (wave efficiency $\eta$) as a function of obstacle density (measured as the ratio of body length and obstacle spacing $L/d$) for nematodes. Error bars represent standard deviations. Error bars represent SDs (n = 26 individuals in open and sparse lattices, n = 20 individuals in the medium lattice, and n = 24 individuals in the dense lattice).}
\end{figure}

We further systematically evaluated nematode locomotor performance in terms of locomotion speed, measured by the wave efficiency $\eta$, the ratio of the forward center of mass speed to the wave propagation speed (Fig. 2B, and refer to Materials and Methods for detailed procedure of wave efficiency measurement). In free swimming, nematodes produced thrust because of the inherent drag anisotropy experienced in a viscous fluid~\cite{sznitman2010propulsive}; for the purposes of this paper we define drag anisotropy as the ratio of the maximum forces on a small element translating through a continuous medium at angles perpendicular and parallel to the element's surface. We noticed that thrust-producing interactions with pillars produce larger $\eta$ relative to the free swimming case~\cite{park2008enhanced,majmudar2012experiments}, despite the similarity of the kinematics. As pillar density was increased, by contrast, wave periodicity was frequently disrupted by inhibitory interactions (or, producing force opposite the direction of travel), typically coinciding with interactions between the nematode's head and a pillar. However, in the densest lattices, bouts of smooth traveling wave propagation between head interactions displayed an overall increase in $\eta$. In this regime, the nematode can take advantage of thrust-producing interactions with the lattice to increase $\eta$ but avoids inhibitory collisions that would lead to jamming and getting stuck. We hypothesized that the mechanism of stabilization is primarily passive in nature, and mechanical intelligence is sufficient for heterogeneity negotiation, without the need of explicit modulations of body postures.

\subsection*{Bilaterally actuated robophysical model development}
To test if mechanical intelligence alone is sufficient to reproduce the performance of nematode lattice traversal, we developed a hard-soft hybrid robophysical model (86 cm long with 7 bending joints) which models the bilateral actuation scheme of nematodes and other limbless organisms, actuating joints by shortening and lengthening cables via decentralized cable-pulley-motor systems (each cable is independently controlled) on either side of each joint (Fig. 1D, Movie S2). By properly coordinating the lengths of cables through waves of angular oscillation passing along the body, this robophysical model can produce similar undulatory locomotion as limbless organisms (Fig. 1B, E). Although its movements are slower than those of limbless organisms, the highly damped nature of the locomotion in both systems (viscous in the nematodes, frictional in the robot) allows the robophysical model to offer insight into the function of mechanical intelligence in complex terrain navigation in the organism. Specifically, we introduce a nondimensional parameter, the ``Coasting number" $\mathcal{C}$, which can be viewed as the ratio of inertial to dissipative forces or as a characteristic timescale for a locomotor to come to rest from steady state speed normalized by a cyclic timescale. For the robot which is dominated by surface friction, $\mathcal{C}$$\sim$$10^{-3}$ whereas nematodes are dominated by viscosity swimming in fluid and $\mathcal{C}$$\sim $$10^{-2}$ (see Supplementary Discussion for calculations of $\mathcal{C}$). To compare the robophysical model and the organism, we assumed they both exist in a regime in which Resistive Force Theory (RFT)~\cite{zhang2014effectiveness,gray1955propulsion} applies with frictional and viscous resistive forces respectively. In this regime, the locomotor performance of a given gait is largely determined by the drag anisotropy and not the specific functional forms of the drag forces (for example, velocity-dependent/viscous versus independent/frictional). In our case, using passive, non-actuated wheels, we experimentally matched the drag anisotropy of the nematodes in the fluid by changing the wheel surface material (refer to Supplementary Methods for a detailed discussion), enabling us to achieve similar performance for nematodes in open fluid and robots locomoting on open, flat terrain.

The bilateral cable actuation mechanism enables body compliance in the robophysical model. However, in contrast to soft limbless robots that inherit compliance from soft materials which are usually hard to modulate, cables in our robophysical model are non-elastic, and thus their lengths can be explicitly controlled. This allows the body compliance in our robophysical model to be quantifiable, programmable, inhomogeneous, and anisotropic, simply by appropriately coordinating the lengthening and shortening of cables. To implement a basic traveling-wave locomotion pattern on the robophysical model as observed in nematodes, we developed the control scheme based on the ``serpenoid" shape-based template~\cite{hirose1993biologically}. The template can generate a central pattern that enables a wave to propagate from head to tail, if the $i$-th joint angle $\alpha_i$ in the spine at time $t$ follows
\begin{equation}
\begin{aligned}
\alpha_i(t) &= A\sin(2\pi\xi \frac{i}{N} - 2\pi\omega t)\\
&= A\cos(2\pi\omega t)\sin(2\pi\xi \frac{i}{N})-A\sin(2\pi\omega t)\cos(2\pi\xi \frac{i}{N})\\
&= w_1(t)\beta_1^\alpha(i) + w_2(t)\beta_2^\alpha(i),
\label{eq:serpenoid}
\end{aligned}
\end{equation}
where $A$, $\xi$ and $\omega$ is the amplitude, the spatial and temporal frequencies of the wave, $i$ is the joint index, and $N$ is the total number of joints. The joint angle $\alpha$ given by this template will be further referred to as the ``suggested" angle (the angle that would be realized absent all external and internal forces apart from those applied by the cables). Thus, the suggested gait path (the trajectory of $w_1(t)$ and $w_2(t)$) forms a perfect circle in the shape space spanned by $w_1$ and $w_2$.

To implement programmable body compliance in the robophysical model, we developed a cable length control scheme based on the suggested angle template, where the lengths of the left and right cables ($L_i^{l}$ and $L_i^{r}$) for the $i$-th joint following
\begin{equation}
\begin{array}{l}
L_i^l(\alpha_i) = \left\{\begin{array}{ll}{\mathcal{L}_i^l(\alpha_i)} & {\text{if } \alpha_i \leq -(2G_i-1)A} \\ {\mathcal{L}_i^l[-A \cdot \min(1, 2G_i-1)]+l_0\cdot[(2G_i-1)A + \alpha_i]} & {\text{if } \alpha_i > -(2G_i-1)A}\end{array}\right. \\ 
L_i^r(\alpha_i) = \left\{\begin{array}{ll}{\mathcal{L}_i^r(\alpha_i)} & {\text{if } \alpha_i \geq (2G_i-1)A} \\ 
 {\mathcal{L}_i^r[A \cdot \min(1, 2G_i-1)]+l_0\cdot[(2G_i-1)A - \alpha_i]} & {\text{if } \alpha_i < (2G_i-1)A}\end{array}\right. \\
\end{array}
\label{eq:policy}
\end{equation}
where $\alpha_i$ is the suggested angle, $A$ is the wave amplitude as in Eq.~\ref{eq:serpenoid}, $\mathcal{L}_i^l$ and $\mathcal{L}_i^r$ are the exact lengths of left and right cables to form $\alpha_i$. $l_0$ is a design parameter that determines how much a cable will be lengthened and is fixed throughout this work (see Supplementary Methods for more discussion). $G_i$ is the generalized compliance for the $i$-th joint, a key controller parameter to enable programmable body compliance. Specifically, in this work we kept the generalized compliance value to be the same throughout all joints, $G_1 = \cdots = G_N = G$. The generalized compliance $G \in [0, +\infty)$ is a parameter that expands the range of possible angles that can occur for a given suggested angle by altering the lengths of the cables on alternate sides; thus $G$ intuitively works as a standalone ``knob" in the control that allows for programmable body compliance -- increasing $G$ leads to more compliance. Moreover, $G$ is a dimensionless quantity that quantifies body compliance and not related to the robophysical model's geometry and characteristics of the environment that the robophysical model locomotes in. 

To provide a better understanding of the generalized compliance $G$, we narrate the robophysical model's compliant states under three representative generalized compliance values below. At $G=0$ the robophysical model is bidirectionally non-compliant (Fig. 3A), where all cables are shortened ($L_i^l(\alpha_i)=\mathcal{L}_i^l(\alpha_i)$ and $L_i^r(\alpha_i)=\mathcal{L}_i^r(\alpha_i)$) so that joints are non-compliant. Note that $\mathcal{L}_i^l(\alpha_i)$ and $\mathcal{L}_i^r(\alpha_i)$ the exact lengths of the left and right cables that are stretched straight to form an angle $\alpha_i$ on the $i$-th joint (see Supplementary Methods for the full deviation of $\mathcal{L}_i^{l}$ and $\mathcal{L}_i^{r}$ based on the robophysical model geometry). When $G=0$, joint angles can precisely track the suggested angles. The projection of joint angle trajectories in the configuration space to the shape space (following the method given by Eq.~\ref{eq:shapeSpaceMap}) then is a perfect circular orbit. Specifically, at $G=0$ the robophysical model behaves as a conventional rigid limbless robot; all joints can resist forces from either sides.

At $G=0.5$ the robophysical model is directionally compliant (Fig. 3B), where either the left or right cable of a joint is lengthened ($L_i^{l}$ or $L_i^{r}$ departs from $\mathcal{L}_i^{l}$ or $\mathcal{L}_i^{r}$) so that the joint is directionally (anisotropically) compliant, thus can admit forces to bend further but reject forces from the other side which would otherwise cause the bend to decrease. In the directionally compliant state, a joint is allowed to form an angle (the emergent angle $\zeta$) with a larger absolute value than the suggested angle ($\alpha$): when a joint is suggested to bend to the right ($\alpha>0$), the left cable will be lengthened (with an amount of $l_0\alpha_i$) so that the joint can be bent further to the right direction, thus its emergent angle $\zeta$ can be larger than the suggested angle $\alpha$, $\zeta \geq \alpha$; and vice versa, the right cable will be lengthened when $\alpha<0$ so that $\zeta \leq \alpha$. Note that when $\alpha=0$, $L_i^{l}(0)=\mathcal{L}_i^{l}(0)$ and $L_i^{r}(0)=\mathcal{L}_i^{r}(0)$ so $\zeta=0$. As a result, the projections of all feasible joint trajectories of $\zeta$ into the shape space yield a feasible region for gait paths to be perturbed by external forces, where the inner boundary is the ``suggested" circular gait orbit. 

At $G=1$ the robophysical model is bidirectionally compliant (Fig. 3C), where both the left and right cables of a joint are lengthened ($L_i^{l}$ and $L_i^{r}$ departs from $\mathcal{L}_i^{l}$ and $\mathcal{L}_i^{r}$) so that the joint is bidirectionally compliant, thus can admit forces from either side. In the bidirectionally compliant state, the emergent angle $\zeta$ of a joint can vary in both directions around $\alpha$: at any given $\alpha$, the left and right are both lengthened (with amounts of $l_0(A+\alpha)$ and $l_0(A-\alpha)$). Note that when $\alpha=A$, $L_i^{r}(A)=\mathcal{L}_i^{r}(A)$ so $\zeta\geq A$, and similarly, when $\alpha=-A$, $L_i^{l}(-A)=\mathcal{L}_i^{l}(-A)$ so $\zeta\leq -A$, meaning the joint will only be directionally compliant when the suggested angle hits its maximum and minimum. In this state, the feasible region of the gait path in the shape space correspondingly expands as the inner boundary shrinks. 

\begin{figure}[H]
\centering
\includegraphics[width=1\textwidth]{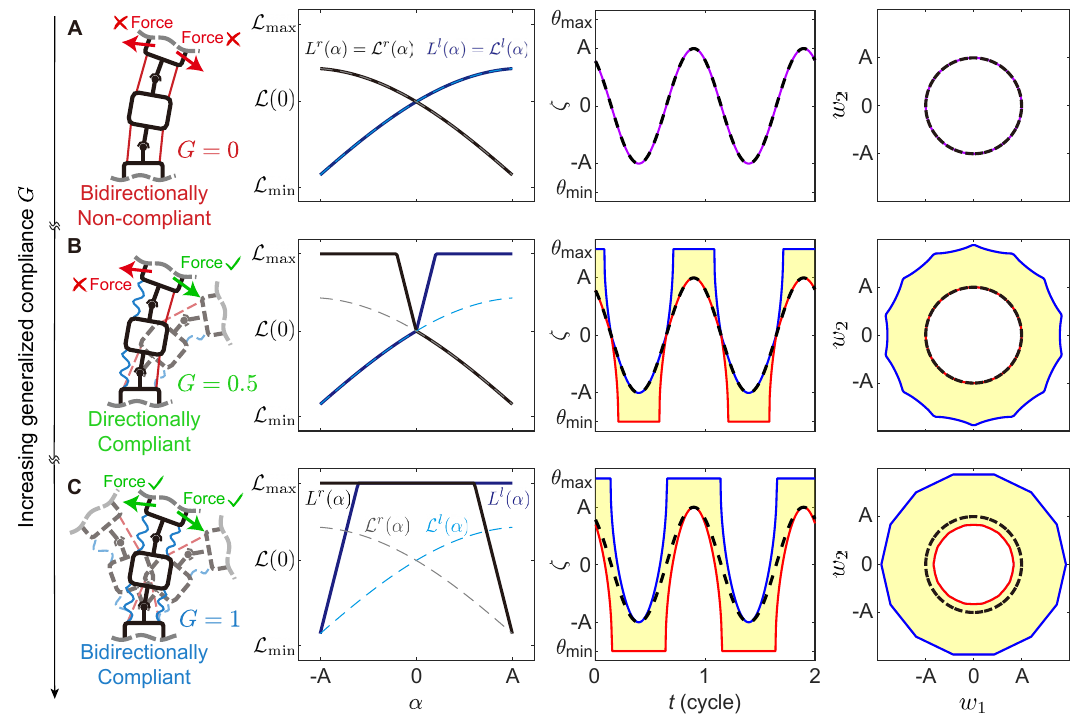}
\caption{Programmable and quantifiable body compliance in the robophysical model. Three representative compliant states of the robophysical model under varied generalized compliance $G$: (\textbf{A}) bidirectionally non-compliant, (\textbf{B}) directionally compliant and (\textbf{C}) bidirectionally compliant. The first column illustrates schematics of cable activation, where red cables are shortened whereas blue cables are lengthened. The second column shows how cables are lengthened at varied suggested angles according to the control scheme, where solid lines represent implemented cable lengths whereas dashed lines represent ``exact" lengths of cables to form the suggested angle. The third column shows how much a feasible emergent angle $\zeta$ (yellow region) is allowed to deviate from the suggested angle $\alpha$ (dashed line), where solid blue and red lines represent upper and lower boundaries of $\zeta$. The last column shows the how much a feasible emergent gait path in the shape space (yellow region) is allowed to deviate from the suggested circular gait path (dashed line), where solid blue and red lines represent outer and inner boundaries of feasible emergent gait paths.}
\end{figure}

As a continuous quantity, when the generalized compliance value falls between representative values described above, the joint can exhibit a hybrid state. For example, when $G=0.75$, the joint will be bidirectionally compliant when $\alpha\in(-0.5A,0.5A)$, and be directionally compliant otherwise. Further, as $G$ value increases passing the bidirectionally compliant representative value, the cable constraints continue to loosen up, until $G$ reaches a point where the joint becomes fully passive. Theoretically, the fully passive value is related to the robophysical model geometry and the gait parameter selection, whereas a consistent value of $1.75$ is observed to correspond with full passivity throughout this work (the full derivation is provided in Supplementary Methods). To sum up, generalized compliance $G$ works as a ``knob" that we tuned to ``program" how strongly the robophysical model is driven by the suggested shape, regulating the level of mechanical intelligence (Movie S2). Thus we varied $G$ in the robophysical model to investigate at which level of mechanical intelligence its locomotor performance can approach nematodes. A full schematic of properties that the robophysical model displays under different $G$ see Fig.~\ref{fig:ConfigPassivity}.

\subsection*{Robophysical model kinematics and performance in heterogeneous terrains}
To test the role of mechanical intelligence in limbless locomotion and its effect on locomotor performance, we conducted robophysical model experiments in four scaled-up environments (from open to dense) corresponding to the nematode study. Similar to the lattices for nematodes, pillars in the lattices for robophysical experiments cannot move and deform upon collision with the robophysical model. In each environment, the robophysical model was under open-loop control, executing a suggested traveling-wave gait as in Eq.~\ref{eq:serpenoid}, with the shape parameters approximated directly from nematode kinematics in the corresponding environment so that the robophysical model used the same gaits as nematodes did (more specifically, the ratio of the body wavelength and the lattice spacing was kept the same between the robophysical model and nematodes, details of the approximation process are provided in Materials and Methods). We varied $G$ to access the locomotion displayed by the robophysical model in each environment. Quantifying locomotor performance (the wave efficiency $\eta$, the ratio of forward center of mass speed to backwards wave propagation speed) across the full range of $G$ revealed that an appropriate $G$ becomes necessary to facilitate open-loop traversal as heterogeneities arise (Fig. 4B). In flat terrain, $\eta$ was inversely correlated to $G$. However, when obstacles were introduced, low $G$ ($\leq 0.5$) resulted in frequent jams and becoming irreversibly stuck. At high $G$ ($\geq 1.5$), the model failed to generate sufficient self-propulsion. $G = 0.75$ emerges as an appropriate $G$ value for locomotion in all heterogeneous environments, as local maxima of $\eta$ display at $G \approx 0.75$ (Movie S3). Further, $\eta$ in the robophysical model with $G = 0.75$ increased as the obstacle density increased, well approaching $\eta$ that displayed in nematodes (Fig. 4C). 

To investigate the emergent robophysical model body kinematics, we tracked emergent joint angles $\zeta$ of the robophysical model, which are comparable to nematode emergent curvatures (detailed reasoning is provided Supplementary Methods). We then projected $\zeta$ onto the shape-basis functions $\beta_{1,2}^\alpha$ to extract the shape space gait path formed by $w_1(t)$ and $w_2(t)$ as we did for nematodes. For $G=0.75$ in the robophysical model, the body kinematics and gait orbits in the shape space (Fig. 4A) closely resembled those observed in nematodes (Fig. 2A). The model performed an approximate traveling wave motion in flat terrain and sparser lattices, which resulted in nearly circular orbits in the shape space. In the dense lattice, analogous to the nematodes, we also observed small deviations from ideal travelling wave shapes, which converged quickly back to the circular orbit. Thus, the robophysical model can serve as an effective model of nematode locomotion, well capturing both overall performance and detailed body kinematics (Movie S4).

\begin{figure}[H]
\centering
\includegraphics[width=0.85\textwidth]{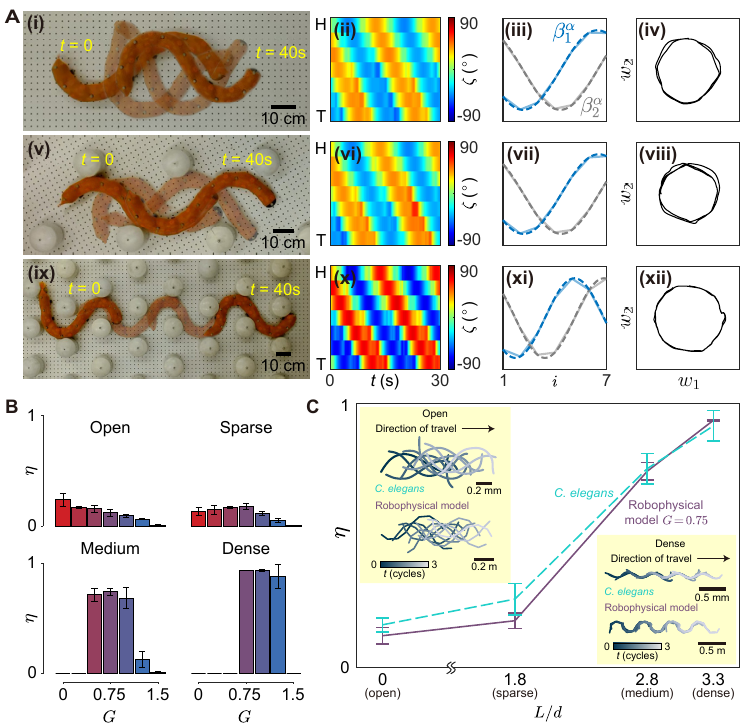}
\caption{Open-loop robot performance reveals the importance of mechanical intelligence. (\textbf{A}) Overlaid snapshots, emergent joint angles, gait paths in the shape space and shape basis of robophysical locomotion ($G=0.75$) in laboratory environments with varied obstacle density. (\textbf{B}) Locomotion speed (wave efficiency $\eta$) of the robophysical model as a function of generalized compliance $G$ in environments with varied obstacle density (open, sparse, medium and dense). Error bars in represent SD across three repetitions per experiment. (\textbf{C}) Comparison of locomotion speed as a function of obstacle density between the biological model \textit{C. elegans} (reproduced from Fig. 2B) and the robophysical model with $G=0.75$, accompanied with example time traces of splined points along the body as the nematode and the robophysical model move in the open and dense environments (insets). Error bars represent the SD across three repetitions per experiment.}
\end{figure}

The emergent match between \textit{C. elegans} and the robophysical model kinematics and the enhancement of performance at $G=0.75$ compared to other $G$ values resulted completely from body compliance -- simply by programmatically and anisotropically loosening the physical constraints on the joints in a way that mirrors the geometry of organismal patterns of activity, which allows joints to passively deform under external forces. Such a seemingly counter-intuitive result (improving performance via relaxing controls) verified our hypothesis that the appropriate level of mechanical intelligence (purely passively, mechanically controlled emergent body-environment interactions) can facilitate heterogeneity navigation, and is sufficient to reproduce organismal lattice traversal performance.

\subsection*{Robophysical model force-deformation characterization}

We used the force-deformation properties of the robophysical model to identify how interactions with obstacles lead to deformations to the suggested traveling wave kinematics that enable successful lattice traversal. By characterizing the relation between the external force $F$ and the emergent joint angle $\zeta$ at suggested angles $\alpha$, we achieved maps of force-deformation properties of the robophysical model with varied $G$ values (Fig. 5, for other $G$ values see Fig.~\ref{fig:ForceCharacterization}). For low $G$, external forces produced minimal deformation of the joint for all parts of the cycle (unless they are sufficiently high to break the cable) (Fig. 5A-i, B-ii). For high $G$, large deformations can be created in response to external forces in either direction (Fig. 5A-iii, B-iii). However, at $G = 0.75$, force-deformation responses displayed a hybrid state (Fig. 5A-ii, B-i): for small angles, force was admitted in both directions (bidirectionally compliant); for large angles, force was admitted in the direction of the bend but stiffly opposed in the opposite direction (directionally compliant).

\begin{figure}[H]
\centering
\includegraphics[width=0.8\textwidth]{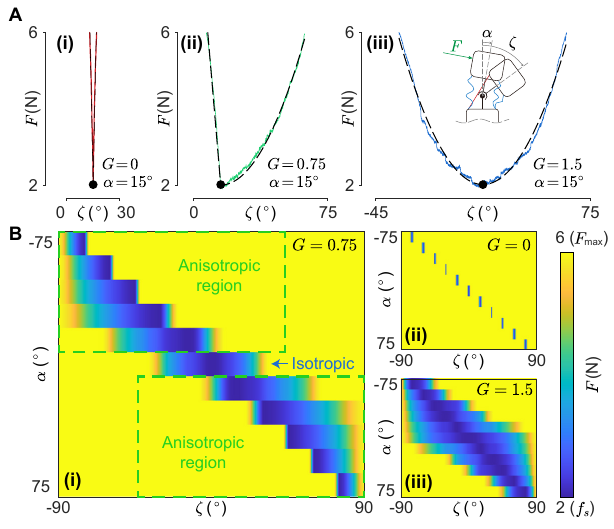}
\caption{Force-deformation characterization for the robophysical model. (\textbf{A}) External force versus emergent joint angle curves that show behaviors of a joint reacting to external forces under different compliance states. (\textbf{B}) Force-deformation maps of the robophysical model with varied $G$ that show the robophysical model body compliance can be programmatically tuned.}
\end{figure}

We hypothesized that such hybrid compliance allows the selective exploitation of thrust-producing interactions through rigid responses and deformations that prevent jamming in detrimental interactions, such as head-on collisions. Our robophysical model and many other limbless undulators move through space by passing body waves from head to tail with wave velocity $v_\text{wave}$ anti-parallel to the center of mass velocity $v_\text{CoM}$ (Fig. 6A). External forces $F_\text{ext}$ from collisions that lie parallel to $v_\text{wave}$ inhibit the center of mass motion, whereas collisions that produce forces parallel to $v_\text{CoM}$ produce thrust. Fig. 6B shows the deflection from the suggested angle in response to a point force ($\approx3$N) parallel or anti-parallel to $v_\text{CoM}$ for a range of suggested joint angles at $G=0.75$. At small suggested joint angles ($|\alpha| < 0.5A$), the joint displays a bidirectional compliant state, in which deflection is permitted more symmetrically ($F_\text{ext}\parallel v_\text{CoM}$ and $F_\text{ext} \parallel v_\text{wave}$) to produce a similar magnitude of deformation. However, as the suggested angle increases ($|\alpha| > 0.5A$), the joint becomes directionally compliant, such asymmetry produces an ``easy" high compliance axis and ``hard" low compliance axis. The direction of the easy and hard axes depends on the shape of the organism. When the ``easy axis" is aligned with inhibitory interactions and the ``hard axis" with thrust producing interactions, organisms can resist buckling while maintaining forward progress. Fig. 6C shows the orientation of the ``easy"/high compliance direction (black triangles) and the ``hard" low compliance (orange triangles) direction for 3 values of $G$ (0, 0.75 and 1.5) and for the various joints along the body of an example 8-link undulator. Small arrows show point forces acting along the body either parallel to $v_\text{CoM}$ or to $v_\text{wave}$. At $G=0$, all joints are non-compliant, hence point forces produce either jamming interactions (small red arrows) or thrust (green red arrows). At $G=0.75$ the distribution of easy and hard axes is arranged such that would-be jamming interactions are converted into body deformations which lead to deflection and therefore successful obstacle traversal, while still maintaining rigidity (non-compliance) in thrust-producing interactions. At $G=1.5$ all interactions permit substantial deformations (all joints are highly bidirectionally compliant). Although jamming is avoided entirely, there is no ability to produce coherent thrust. Experimentally, the geometry of contacts closely follows the curvature profile of the gait (Fig.~\ref{fig:ContactMap}). Would-be jamming interactions, for example near the head, often lead to longer durations of contact, governed by the dynamics of the deformation under locally compliant joints, whereas thrust-producing interactions at higher curvature near the mid-body typically follow regular contact patterns with shorter duration contacts, matching the propagation of curvature along the body.

\begin{figure}[H]
\centering
\includegraphics[width=0.8\textwidth]{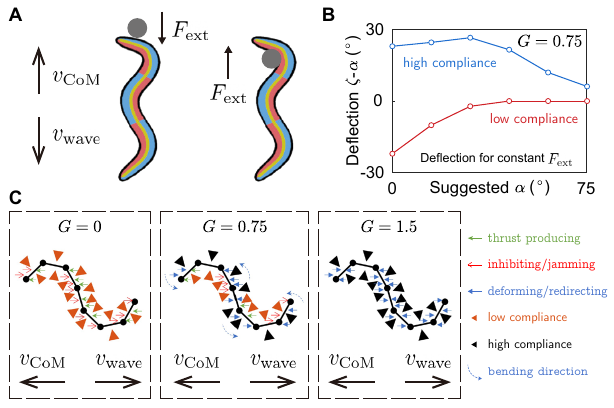}
\caption{A simplified model to understand the functional mechanism of mechanical intelligence. (\textbf{A}) Schematic illustration of an undulator facing inhibitory interactions (left) and thrust producing interactions (right). (\textbf{B}) Deflection angle in response to a point force $F_\text{ext}$ either parallel or anti-parallel to $v_\text{CoM}$ at $G = 0.75$ for different commanded angles, showing the response of the ``easy" or high compliance direction and the ``hard" low compliance direction. (\textbf{C}) The geometry of easy (black triangles) and hard directions (orange triangles) for a single posture across three representative values of $G$. Small arrows show point forces that are thrust producing (green arrows), are jamming (red arrows), or result in deformation of the undulator from the commanded shape (blue arrows), with bend directions indicated by the dashed blue lines.}
\end{figure}

This simplified model (Fig. 6) revealed that for certain, intermediate values, of $G$, the robophysical model spontaneously converted inhibitory interactions into soft deflections while maintaining rigidity and thrust production in advantageous collisions without any explicit computation. The coordinated shortening and lengthening of the cables served therefore not only to realize an approximate traveling wave body shape sequence, but also to dynamically modulate the compliance properties of the robot to buffer the motion to external collisions.

\subsection*{Emergent head behaviors in nematodes and the robophysical model}
The robophysical model displayed emergent functional behaviors when $G=0.75$. Upon collision in the head, two typical head interaction events emerged in the robophysical model to exploit the asymmetric force-deformation response: ``gliding" where the head slides near-tangentially past the obstacle (Fig. 7A-i), and ``buckling" where a collision induces a momentary increase in the local curvature near the head, which then facilitates a shallower angle of attack (Fig. 7A-iii). Gliding led to only minor deviations from circular paths in the shape space (Fig. 7A-ii), whereas buckling led to larger deviations from the circular orbit, as the radius of the path increased at constant phase angle (Fig. 7A-iv). This transient cessation of the wave phase velocity arose as the obstacle restricted the forward progress, constraining the body and inducing increased curvature. Among all the events we collected ($n \approx 100$), we classified $33.6\%$ as buckling (with a phase pause over 0.5 s) and other $66.4\%$ as gliding. Given such behaviors took place in the open-loop robophysical model only commanded with a suggested traveling retrograde wave, the gliding and buckling behaviors instigated by collisions occurred passively, and therefore were dominantly determined by passive body-environment interactions.

Given the correspondence on gross locomotor performance and body kinematics of the robophysical model and \textit{C. elegans}, and the importance of head gliding and buckling dynamics in facilitating lattice transport, we next investigated if \textit{C. elegans} displayed similar head (or neck) dynamics during obstacle interactions. We observed substantially analogous behaviors (Fig. 7B-i to B-iv, Movie S5) such that 28.6\% of head interaction events were classified as buckling (with a phase pause over 0.2s) whereas the rest were considered gliding ($n \approx 100$). We thus posit that the nematodes' head interactions help passively facilitate locomotion in heterogeneous environments, as manifestations of mechanical intelligence. Specifically, potentially inhibitory collisions that might lead to jamming can be mitigated by the asymmetrical compliance in the head. 

\subsection*{Active reversals in nematodes and the robophysical model}
Other than ``gliding" and ``buckling" head events, we also noted that in some instances \textit{C. elegans} displayed a ``reversal" behavior (Fig. 7B-v) correlated with collisions which we did not see in the open-loop robophysical model. The reversal behavior is an actively controlled behavior~\cite{zhao2003reversal}, in which nematodes initiate a reversal of the direction of the traveling wave for a short period, and then repropagate the original traveling wave (Fig. 7B-vi). We hypothesized that active responses to heterogeneities (even as simple as reversals induced by head collisions) could benefit locomotion by augmenting mechanical intelligence. The active reversals induced by high angle of incident collisions supplement mechanical intelligence by providing an alternative means of modulating the angle of attack. This reversal behavior is likely initiated by mechanosensory neurons in the head, such as FLP (Fig. 7C) which have stereotyped anterior cellular processes which likely transduce mechanical inputs into signals that produce the reversals~\cite{doroquez2014high}. 

Similar to theoretical and computational models in biomechanics, robophysical models allow tests of hypotheses that are inconvenient with living systems. Thus, we next used the robophysical model to probe possible functional locomotor roles of the active reversal behaviors positing that the inherent mechanical intelligence in the nematode could be augmented by simple head collision sensing feedback. To do so, we developed a head collision sensor (a force-sensitive resistor array, manufacturing and control details given in Marterials and Methods) for the model (Fig. 7C) to allow real-time collision angle and force estimation. To realize reversal behavior, we programmed the device to reverse the direction of wave propagation when a harsh head collision (large collision force and angle) was detected. 

\begin{figure}[H]
\centering
\includegraphics[width=0.8\textwidth]{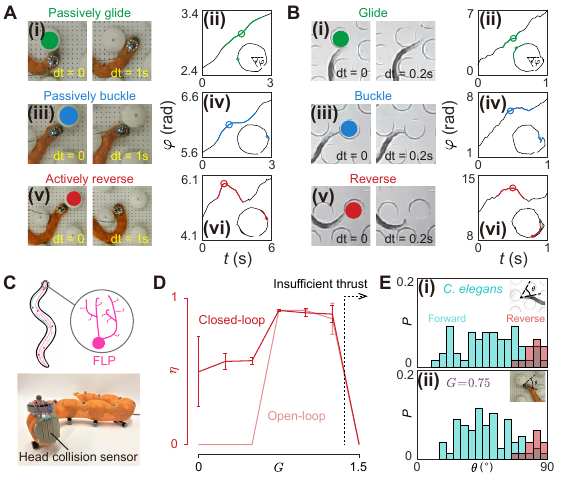}
\caption{Mechanical intelligence enables passive behaviors, and can be augmented actively. (\textbf{A}) The passive ``gliding" and ``buckling" behaviors, and an active ``reversing" behavior in the robophysical model, along with their corresponding characteristic phase-time plots. (\textbf{B}) Analogous behaviors displayed by the biological model, along with corresponding phase-time plots. (\textbf{C}) The FLP dendrite sensory structure in nematodes, and the head collision sensor in the robophysical model for studying how reversals augment mechanical intelligence. (\textbf{D}) Wave efficiency as a function of $G$ in the dense environment for the robophysical model with and without reversals, showing reversals can robustify robophysical locomotion. Error bars represent SDs across three repetitive trials of each experiment. (\textbf{E}) Head collision angle probability distributions classified by post-collision motion directions (forward or reverse) in nematodes and the closed-loop robophysical model ($G=0.75$).}
\end{figure}

We studied the closed-loop robophysical model with reversal capability in the dense environment and compared its locomotor performance to open-loop results. Reversals enabled the robophysical model to traverse the environment in the low generalized compliance regime, which the open-loop strategy failed to (Fig. 7D, Movie S6), improving $\eta$ in the range $0 \leq G \leq 0.5$. The reversal behaviors robustified the locomotion by increasing the range of $G$ that allows the model to effectively locomote in the most challenging environment. The closed-loop robophysical model also showed substantially similar kinematics as observed in nematodes (Fig. 7A-v, vi, and Fig. 7B-v, vi, Movie S5). Robophysical experiments revealed the function of reversal behaviors in undulatory locomotors: by not simply repeating the same movement back and forth in place, reversals allow the locomotor to take advantage of mechanically intelligent dynamics -- passively adjusting body postures and spontaneously finding favorable position and orientation to generate effective thrust for locomoting further. 

Given the similarity in behavioral kinematics between the closed-loop robophysical model and nematodes, we further investigated head collision angles and corresponding post-collision movement directions (forward or reverse) in both systems. The probability distributions of head collision angle for forward and reverse motion further demonstrate that the reversal-capable robophysical model with $G=0.75$ can well capture emergent behaviors that are induced by mechanical intelligence in \textit{C. elegans} (Fig. 7E, and probability distributions for other $G$ values are shown in Fig.~\ref{fig:HeadProb}), thus works as a reliable model of \textit{C. elegans} locomotion (an example comparison of body kinematics see Fig.~\ref{fig:RobotWormComp}, also note that this result applies to the presented robophysical model design and controls, given the robophysical model's reversal behavior can be altered by a different head sensor implementation). Such qualitative agreement in body kinematics and the quantitative agreement in body event statistics imply that simple computational intelligence (reversals triggered by head sensing feedback) can compensate the short of mechanical intelligence (especially at the low-$G$ region), or enhance mechanical intelligence (in terms of introducing extra chances for passive body-environment interactions), thus can augment locomotor performance. This also provides insight into the functional mechanism of the seemingly inefficient reversal behaviors displayed in nematodes. Our results also suggest that the spatiotemporal responses of the head sensory neurons such as FLP~\cite{chatzigeorgiou2011lateral} may be tuned to help facilitate obstacle navigation. For instance, the spatial structure of the cellular processes within the head (Fig. 7C) may allow the nematode to sense the collision angle, explaining the angular dependence of the different collision behaviors (Fig. 7E). Further, the robophysical model demonstrates as a comprehensive example of embodied intelligence~\cite{pfeifer2007self,iida2022timescales} and morphological computation~\cite{hauser2011towards,laschi2016lessons}, displaying the most robust locomotion capabilities while working under the synergies of mechanical intelligence and computational intelligence. 

\subsection*{Open-loop robot capabilities in laboratory complex environments}

Nematodes not only perform well in heterogeneous, collision-dominated environments. They also encounter a diverse array of substrates, including Newtonian fluids of varying viscosity and other flowable substances with complex, non-Newtonian rheologies~\cite{felix2010natural}. Hence, body compliance that enables lattice traversal, may also improve performance in less structured environments or, at a minimum not disrupt performance. We, therefore, hypothesized that our bilaterally actuated limbless robophysical model would also display good performance without major changes in control in a diversity of robophysical model terrain with properties similar to those encountered during search and rescue and other applications. Indeed, we found that beyond functioning as a model for discovering and understanding emergent principles in limbless locomotion that cannot be directly tested with organisms, the bilaterally actuated limbless robot displayed substantial terrestrial mobility in diverse, complex, and more challenging environments. 

We tested the robot in a range of laboratory and outdoor environments (Fig. 8, Fig.~\ref{fig:RobotDemos}, Movie S7 and Movie S8). Beyond regular lattices, the robot demonstrated effective traversal in randomly distributed obstacle terrains (Fig.~\ref{fig:RobotDemos}A) and agile transitions from open terrain to obstacle terrain (Fig.~\ref{fig:RobotDemos}B), where the robot was under open-loop controls with $G=0.75$. Without the need for active adaptation of body shapes~\cite{wang2020directional,kano2013obstacles,travers2016shape} or selection of paths~\cite{sartoretti2021autonomous,bing2020perception,hanssen2020path} based on the awareness of internal states (such as instantaneous joint angles or torques) or knowledge of the surrounding environment (for example, via contact sensing or visual feedback) as proposed in previous works, the mechanical intelligence in this robot enables compliant body-environment interactions, facilitating the spontaneous locomotion. 

Further, we conducted tests of locomotion speeds and cost of transport in other types of environment, firstly granular media (Fig.~\ref{fig:RobotDemos}C), a model flowable medium previously studied using other limbless systems~\cite{maladen2011undulatory,chong2023gait}. In the granular material, we found that introducing an appropriate amount of passive body mechanics (by increasing $G$) can substantially reduce energy consumption without a notable loss of locomotor speed, with the local minimum in the cost of transport emerging at intermediate values of $G$ (Fig.~\ref{fig:RobotCOT}). 

We also tested the robot in narrow channels which function as models of pipes (Fig.~\ref{fig:RobotDemos}D), where we reversed the direction of the wave propagation to enable forward locomotion on the robot without wheels relying purely on wall interactions (see Supplementary Discussion for more). The generalized compliance $G$ enables spontaneous shape adaptation/modulation to a highly constrained channel without the need of probing the channel width in advance. Further, the local minimum of cost of transport emerged at high compliance $G=1.25$. We also measured cost of transport in lattices with varied obstacle density (sparse, medium and dense as discussed previously), where we found local maxima of locomotion speed and local minima of cost of transport all emerged at intermediate values of $G$. More detailed robot performance results and further discussions are included in Supplementary Discussion.

\subsection*{Open-loop robot capabilities in natural complex environments}

To determine the potential benefits of mechanical intelligence in practical limbless robot applications and the generalizability of principles derived from two-dimensional laboratory environments to complex three-dimensional natural environments, we conducted open-loop locomotion experiments in a mechanically complex environment. Specifically, we tested the robot's performance in a randomly distributed and tightly packed pile of rocks (Fig. 8 and Movie S8), simulating the terradynamic challenges that a limbless robot may face during search-and-rescue or planetary exploration tasks. Our quantitative analysis of robot locomotion performance demonstrated that, with an appropriate amount of generalized compliance ($G=0.75$), mechanical intelligence facilitates effective negotiation with irregularities, ensuring successful locomotion. Conversely, inadequate compliance ($G=0$) hindered obstacle traversal, whereas excessive compliance ($G=1.5$) resulted in insufficient thrust generation. Notably, the cost of transport exhibited local minima at intermediate values of $G$, consistent with our findings from laboratory tests.

Overall, laboratory and outdoor tests demonstrated that intermediate values of $G$ enable effective locomotion in the largest range of environments and provide reduced costs of transport. This suggests mechanical intelligence not only facilitates obstacle negotiation, but also can improve locomotion speed and efficiency. 

\begin{figure}[H]
\centering
\includegraphics[width=0.6\textwidth]{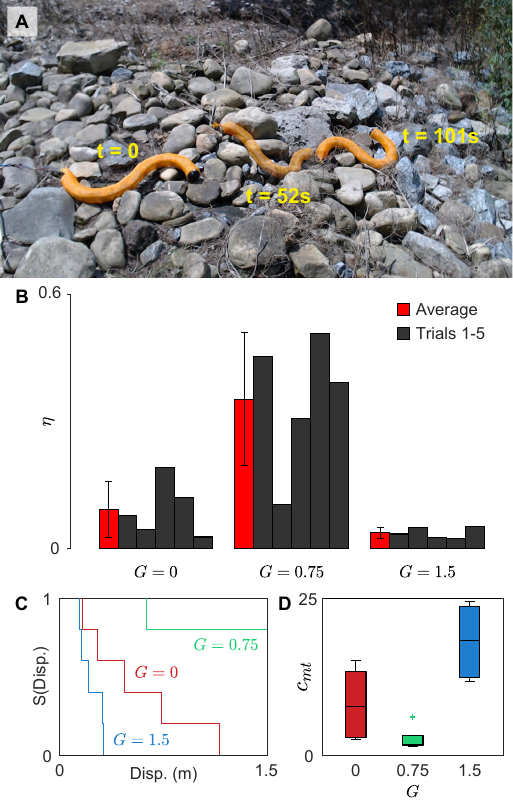}
\caption{Open-loop robot capabilities in real-world complex environments. (\textbf{A}) Time-lapse photos of the open-loop robot traversing over a tightly packed rock pile with an intermediate generalized compliance value ($G=0.75$). (\textbf{B}) Comparison of locomotion speed (wave efficiency $\eta$) with varied $G$ on the rock pile. Error bars represent SDs. (\textbf{C}) The survivor function for varied $G$ with respect to displacement, measuring the robot's traveling distance before getting stuck or failing in motors. (\textbf{D}) Mechanical cost of transport ($\mcot$) for varied $G$ on the rock pile, measuring the robot's energy efficiency of locomotion. Box central mark indicates the median, edges indicate the 25-th and 75-th percentiles. The whiskers cover data points within a range of 1.5 times the interquartile range, whereas outliers outside of this range are marked with a + symbol.}
\end{figure}

\section*{Discussion}
In summary, our integrative and comparative study of biological and robophysical limbless locomotors reveals that mechanical intelligence, the general collection of emergent adaptive behaviors that arise from passive body-environment interactions, simplifies control in terrestrial limbless locomotion, especially in heterogeneous environments, and is sufficient to reproduce organismal lattice traversal performance. The robophysical model, once programmed with an appropriate level of compliance, not only accurately models undulatory organisms in terms of locomotor performance and body kinematics, but also in terms of dynamic force-deformation relationships (similar force-deformation relationships have been established for vertebrate undulators~\cite{long1998muscles}). Dynamic force-deformation relationships are non-trivial for an organism of the scale of \textit{C. elegans} (only passive viscoelastic properties have been determined for \textit{C. elegans}~\cite{backholm2013viscoelastic}). Thus, our robophysical model is a useful tool for understanding the functional mechanism of mechanical intelligence in the organism -- by identifying and understanding the mechanically intelligent control regimes of the robophysical model that accurately reproduce \textit{C. elegans} kinematics in lattices, we can generate hypotheses about what underlying physiological and anatomical details are required to produce the emergent effective locomotion. Broadly, model organisms like \textit{C. elegans} have an important role to play in connecting neural dynamics to behavior. Our results suggest that mechanics also play a substantial role in shaping behavior via processes that occur outside the nervous system, and therefore must be understood and accounted for to reach a comprehensive understanding of animal behavior in general.

Robotic limbless locomotion in confined environments presents challenges in generating adequate thrust and preventing jamming caused by obstacles. Prior research has confronted this challenge through gait design and online parameter turning approaches~\cite{transeth2008snake,sanfilippo2017perception,wang2020directional,chong2023gait}. Essentially, if provided with sufficient foreknowledge of the environment or precise real-time proprioceptive sensory feedback (such as visual or internal body forces), it is possible that an ``optimal" gait template can be carefully designed, or ``optimal" parameters within a template can be tuned online so that even a non-compliant robot can move effectively. In the case of lattices, optimal gaits will have wavelengths, amplitudes and phasing that allow geometric conformity to the lattice (in other words, the wavelength and lateral displacement, determined by the amplitude, will be an integer multiple of the lattice spacing). However, developing and implementing such controllers and sensing modalities requires considerable effort and computational resources. Our approach of exploiting mechanical intelligence can replace these complicated processes, enabling the robot to move in complex environments with open-loop controls, utilizing a simple traveling wave template with low sensitivity to the chosen wave parameters (so that slightly mismatched parameters do not fail to produce locomotion because of mechanical modulation of commanded shapes). Further, we verified in laboratory and natural complex environments that mechanical intelligence (in the form of the appropriate compliant actuation scheme) can even improve locomotion speed and efficiency. For nematodes, who rely on mechanical and chemical cues to navigate, gait selection based on foreknowledge of the environment is not possible; hence the mechanical control scheme is likely important in traversing dense terrain. Even in organisms with vision, like snakes, the speed of locomotion often makes gait planning ineffective, and passive mechanisms again become substantial~\cite{schiebel2019mechanical}. Moreover, our results hint at mechanisms that govern the trade-off between active neural controls and passive body mechanics in nematodes. Our comparative exploration of mechanical intelligence could potentially offer a perspective that complements existing approaches to the question of the general role of neural versus mechanical control~\cite{gazzola2015gait,boyle2012gait,denham2018signatures,haspel2021resilience,majmudar2012experiments}.

Further, our demonstration of the advantages arising from our implementation of mechanical intelligence through the bilateral actuation mechanism presents several promising research avenues. As we observed in experiments that the performance of the robophysical model operating at a certain $G$ value can vary in different environments, we posit that developing a full mechanistic model of the dynamics of the system in various environments could further help determine ``optimal" $G$ based on terrain properties. As $G$ can be dynamically tuned, we posit that adding sensory capabilities could enable the robot to learn or select the ``optimal" $G$ value in real-time that accommodate best with the current environment. As each joint is controlled in a decentralized manner, we posit that locally varying $G$ based on local sensing feedback would enable the system to maximize the utility of surrounding environment to generate thrust and thus to locomote more effectively.

Finally, the bilateral actuation scheme suggests a design and control paradigm for limbless robots. Contrasting the lack of mechanical intelligence in limbless robots to date, the bilateral actuation mechanism offloads complex sensorimotor controls for handling body-environment interactions to mechanical intelligence, improving locomotion efficiency and freeing up onboard hardware and computational bandwidth for advanced sensing and motion planning techniques~\cite{sanfilippo2017perception,sartoretti2021autonomous,wang2022generalized,chong2021frequency,takemori2022adaptive,ijspeert2008central,thandiackal2021emergence,bellegarda2022cpg,haarnoja2017reinforcement,ramasamy2021optimal}. This represents a paradigm shift in limbless robotics that could pave the way for the future development of more agile, intelligent and capable limbless robots that fulfill their promised potential of maneuverability in extremely complex environments, finding diverse applications such as search and rescue, industrial inspection, agricultural management, and planetary exploration. 

\section*{Materials and Methods}

\subsection*{Biological experiments and data processing}
Wild-type \textit{C. elegans} (QLN2) was used for all experiments. Nematodes were cultured using standard protocols on NGM agar plates with \textit{Escherichia coli} (OP-50) lawns. Nematodes were cultured at $20^{\circ}$C and synchronized to day-1 adults for all studies. 

Sequences of body curvatures over time of nematode locomotion in lattice were extracted from video recordings (details of lattice manufacturing and body curvature extraction are provided in Supplementary Methods). To simplify the analysis of locomotion kinematics in lattices, we exploit dimensionality reduction techniques. Prior work illustrated that the majority of body postures in undulating systems can be described by linear combinations of sine-like shape-basis functions, despite the inherently high dimensionality of postural data~\cite{rieser2019geometric,chong2022coordinating}. We assume that the essence of the body curvature profile $\kappa$ at time $t$ and location $s$ ($s=0$ denotes head and $s=1$ denotes tail) can be approximated by Eq.~\ref{eq:shapeSpaceMap}, where $\xi$ is the spatial frequency of body undulation obtained from direct fitting. $w_1(t)$ and $w_2(t)$ are the reduced shape variables describing the instantaneous shape of the locomotor at time $t$. Thus, the locomotion may be visualized as a path through a two-dimensional ``shape space" defined by $w_1$ and $w_2$. Practically, we first performed principal component analysis to the curvature data ($\kappa(s,t)$) to extract the first two principle components, which account for over 90\% of the variation in observed body configurations (Fig.~\ref{fig:PCs}). Then we fitted two shape-basis functions, in the form of $\sin(2\pi\xi s+\phi)$ and $\cos(2\pi\xi s+\phi)$, to the principle components (examples shown in Fig. 2A-iii, vii, xi, where $\xi=0.81, 0.80, 1.75$ for presented examples, respectively). We projected the curvatures onto the shape-basis functions, by finding the least-squares solution~\cite{strang1993introduction}, to extract the weights of shape-basis functions, reduced shape variables $w_1(t)$ and $w_2(t)$. The gait path then is the trajectory formed by $w_1(t)$ and $w_2(t)$ in the shape space spanned by $w_1$ and $w_2$. 

Collision events with pillars were identified manually, and the angle of the head and the pillar were calculated manually in ImageJ. To calculate wave efficiencies, bouts of locomotion containing at least 3 cycles of forward movement were selected. The wave efficiency is calculated as $\eta = v_\text{CoM}/v_\text{wave}$, where $v_\text{CoM}$ is the center of mass velocity of the organism and $v_\text{wave}$ is the wave speed. $v_\text{CoM}$ was calculated directly from microscopy videos using the distance traveled by the nematode's head over an integer number of wave cycles. The wave speed $v_\text{wave} = f\lambda$ was calculated using the measured frequency and wavelength of each nematode. For the head collision angle of nematodes, we measured the angle between the body centerline and the tangential line of the pillar that passes the contact point. To classify gliding and buckling in the collection of head collision events which followed with forward body movement (no reversal), we examined the nematode body kinematics and calculated phase over time around the collision time. We classified the events that led to a phase pause over 0.2 second as buckling and others as gliding. 

\subsection*{Robophysical experiments}
We built laboratory models of heterogeneous terrains (Fig.~\ref{fig:Terrains}B) scaled to the dimensions of the robot, comparable to those used in biological experiments. The wheels coated by low-friction fiberglass tape that were equipped on the robophysical model can create a $\sim$1.6:1 drag anisotropy, which is close to that for nematode in the liquid between pillars, assumed to be modelled by a cylindrical cross section in a low Reynolds number viscous fluid~\cite{gray1955propulsion,sznitman2010propulsive}. Note that, the magnitude of reaction force on wheels of the robophysical model is speed independent~\cite{rieser2019dynamics}, whereas the magnitude of reaction force is linearly dependent on speed for nematodes in viscous fluid. However, drag anisotropy is the dominant factor in governing performance in undulatory locomotion~\cite{alben2019efficient}, and the difference between frictional and viscous drag are likely to be subtle. 

Finally, given that nematodes displayed different gaits in lattices with different densities, the robophysical model's suggested gait must be selected to replicate the kinematics of nematode locomotion. In each corresponding environment, we kept the ratio of the wavelength displayed on the body and the spacing of pillars in the lattice the same between the robophysical model and nematodes. This ensures that the robophysical model and the nematodes have similar periodic contacts with the lattice (Fig.~\ref{fig:ContactMap}). As described previously, we tracked the centerlines of the nematode body in video recordings and approximated the wavelength of the nematode posture in each frame. We then averaged the wavelengths for all the frames and divided them by the pillar spacing of the lattice, yielding the wavelength-spacing ratio ($\sim$2 for the sparse environment, $\sim$2.2 for the medium environment, and $\sim$1.8 for the dense environment). On the robophysical model, we tuned the amplitude $A$ and spatial frequency $\xi$ in the suggested gait in Eq.~\ref{eq:serpenoid} so that the robophysical wavelength-spacing ratio matches with nematode in each scaled-up environment. Specifically, in this work, we used $A = 46^\circ, 48^\circ, 51^\circ, 72^\circ$ and $\xi = 0.82, 0.80, 0.58, 1.02$ for open, sparse, medium, and dense environments, respectively. However, note that the choices of these parameters depend on the robophysical model's dimensional specifications, such as module length and the maximum range of joint bending. 

\subsubsection*{Environment setup}
The robophysical experiments were conducted on a level pegboard (The Home Depot) measuring 2.4 meters in length and 1.2 meters in width, with 6.35-mm holes spaced at every 25.4 mm. Each hole has screw inserts that are fitted for 4-mm bolts that can be used to secure PVC pipe caps. The pipe caps (Charlotte, 12.7 cm in diameter and 10 cm in height) were used as reconfigurable obstacles in the experiments. They have 4-mm holes drilled at their center and can be secured to the pegboard using long 4-mm bolts (McMaster-Carr) that were fastened into the screw inserts. An example lattice configuration is shown in Fig.~\ref{fig:Terrains}B. This experimental setup allows for obstacles to be easily rearranged and spaced on the pegboard to match the pillar spacings of different lattices in the nematode experiments.

The OptiTrack motion-tracking system was utilized to record the positions and postures of the robophysical model in the workspace. Six IR cameras (OptiTrack Flex 3) were mounted above the lattice to capture the real-time 3D positions of nine reflective markers attached to the robophysical model's body, including seven at each joint, one at the anterior end, and one at the posterior end. The X, Y, and Z position values of each marker were obtained from the Motive software using MATLAB. In addition, a high-resolution camera (Logitech HD Pro Webcam C920) was mounted above the experiment environment to record videos of each experiment. The footage was used to analyze the head collision angle probability distributions classified by post-collision motion directions.

\subsubsection*{Experiment and analysis}
Robophysical experiments consisted of a series of trials running the robophysical model in the lattices. One trial was running the robophysical model from an initial position until it reached one of the following states: 1) the robophysical model exited the lattice; 2) the robophysical model got stuck (did not proceed for 10 consecutive gait cycles); or 3) any of the servo motors overloaded (experiencing torque that exceeded the stall torque). Three separate trials were conducted for each generalized compliance value (ranging from $0$ to $1.5$ with a $0.25$ interval) in each of the four environments (open, sparse, medium, and dense). To ensure consistency across trials, three initial positions were randomly selected and kept identical for all values of generalized compliance. 

For the analysis of the robophysical model kinematics, we extracted emergent joint angles $\zeta$ using tracked positions of the markers. Given that we view the joint angles in the discretized body equivalent to the curvatures in the continuous body, similar to nematodes, we projected $\zeta$ (that can vary with $G$) onto the suggested shape-basis functions $\beta_{1,2}^\alpha$ (that remain the same for all $G$'s in a specific lattice setup) as in Eq.~\ref{eq:serpenoid}, by finding the least-squares solution. This allows us to extract the reduced shape variables $w_1(t)$ and $w_2(t)$ and to analyze the robophysical model's emergent gait paths in the shape space. The methods for the calculation of wave efficiency and the measurement of head collision angle in the robophysical model are the same as nematodes, based on tracked data. The method for classification of the passive behaviors is the same as well, whereas the threshold of phase pauses for buckling classification was 0.5 second for the robophysical model. 

\subsection*{Force-deformation characterization experiment}
Force-deformation experiments were performed by measuring the relation between the magnitude of an external pushing force exerted on an joint with a certain $G$ and the emergent joint angle. We designed a custom, 3D printed stick to push the robophysical model. The stick was attached to a load cell (FUTEK LLB350-FSH03999), and the load cell was mounted the force sensor to a robot arm (DENSO VS-087), as shown in Fig.~\ref{fig:ForceCharacterization}A. The robot arm was programmed to move the stick in a circular trajectory at a constant 1 mm/s velocity, where the center of the circle was colinear with the rotation axis of the joint and the radius of the circle was 60 mm such that the pushing point was at the middle of the module. Analog signals of the load cell were passed through an analog amplifier (FUTEK IAA100), then an analog to digital multifunctional data acquisition module (NI USB-6009), and the digital signal was recorded using NI LabVIEW. The robot body was fixed to a rigid table using two wooden planks that are firmly secured to the table. The robot is pinched between the planks, fixing it to the table surface. One single robot joint is left extending out past the planks for the force-deformation experiments. The joint is given a specified joint angle and $G$ value before the start of the experiment. Specified angle values were swept from -75$^\circ$ to 75$^\circ$ with an increment of 15$^\circ$.

In each experiment, the end effector of the robot arm began rotating in the clockwise direction from the suggested angle until the force reaches a maximum value, set as 6 N which is sufficiently large to bend a compliant joint but would not break the robophysical model. And this process was repeated in the counter-clockwise direction starting from the suggested angle. Taking all force-emergent angle characterizations together, we show maps of force-deformation properties with varied $G$ (Fig~\ref{fig:ForceCharacterization}B). 

\section*{Supplementary Materials}
\noindent Supplementary Methods

\noindent Supplementary Discussion

\noindent Figs. S1 to S12

\noindent Captions for Movies S1 to S8

\bibliographystyle{Science}
\bibliography{mainbib}

\section*{Acknowledgments}
The authors would like to thank Zhexin Shen for the help with manufacturing the head collision sensor, Daniel Soto for the help with construction of robophysical experiment environment, Valerie Zborovsky for the help with robphysical (indoor) experiments, Zhaochen Xu, Anushka Bhumkar, and Nishanth Mankame for their help with robphysical (outdoor) experiments, Zach James for the help with biological experiments, and Lucinda Peng for useful discussions. 

\noindent \textbf{Funding:} The authors are grateful for funding from Army Research Office Grant (W911NF-11-1-0514), National Science Foundation Physics of Living Systems Student Research Network (GR10003305), NSF-Simons Southeast Center for Mathematics and Biology (National Science Foundation DMS1764406, Simons Foundation SFARI 594594), and the Dunn Family Professorship. 

\noindent \textbf{Author contributions:} 

\noindent Conceptualization: TW, CJP, VHK, DIG

\noindent Methodology: TW, CJP, VHK, BC, KD

\noindent Investigation: TW, CJP, VHK, BC, KD

\noindent Visualization: TW, CJP

\noindent Funding acquisition: DIG, HL

\noindent Project administration: DIG, HL

\noindent Supervision: DIG, HL

\noindent Writing -- original draft: TW, CJP, DIG

\noindent Writing -- review \& editing: TW, CJP, VHK, BC, KD, HL, DIG

\noindent \textbf{Competing interests}: Some of the subject matter herein may be implicated in one or more pending patent applications such as PCT Patent Application No. PCT/ US2022/043362, entitled “DEVICES AND SYSTEMS FOR LOCOMOTING DIVERSE TERRAIN AND METHODS OF USE”, which claims the benefit of priority to US Provisional App. No. 63/ 243,435 filed 09/13/2021, and US Provisional App. No. 63/ 318,868 filed 03/11/2022. The authors declare that they have no other competing interests.

\noindent \textbf{Data and materials availability}: All data are available in the main text or the supplementary materials. 

\newpage
\setcounter{page}{1}
\resetlinenumber
\title{Supplementary Materials for \\
Mechanical Intelligence Simplifies Control in Terrestrial Limbless Locomotion}


\author
{Tianyu Wang \textit{et al.}
\\
\normalsize{$^\ast$Corresponding author: Daniel I. Goldman, daniel.goldman@physics.gatech.edu}\\
}


\date{}


\baselineskip24pt


\maketitle 
\setcounter{figure}{0}
\renewcommand{\figurename}{Fig.}
\renewcommand{\thefigure}{S\arabic{figure}}
\setcounter{equation}{0}
\renewcommand{\theequation}{S\arabic{equation}}
\setcounter{section}{0}
\renewcommand{\thesection}{S\arabic{section}}

\noindent \textbf{The PDF file includes:}

\noindent Supplementary Methods

\noindent Supplementary Discussion

\noindent Figs. S1 to S12

\noindent Captions for Movies S1 to S8

\noindent \textbf{Other Supplementary Material for this manuscript includes the following:}

\noindent Movies S1 to S8

\newpage
\section*{Supplementary Methods}

\subsection*{Biological experiment}
\subsubsection*{Lattice setup}
\begin{figure}[H]
\centering
\includegraphics[width=0.9\textwidth]{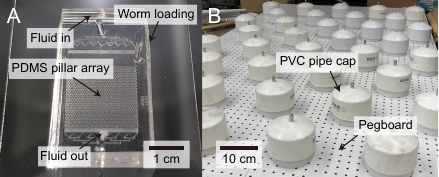}
\caption{Heterogeneous environments for investigating mechanical intelligence in limbless locomotors. (\textbf{A}) A microscopic pillar array for studying locomotion of \textit{C. elegans}. (\textbf{B}) A macroscopic obstacle terrain for studying locomotion of the robophysical model.}
\label{fig:Terrains}
\end{figure}

\begin{figure}[H]
\centering
\includegraphics[width=0.9\textwidth]{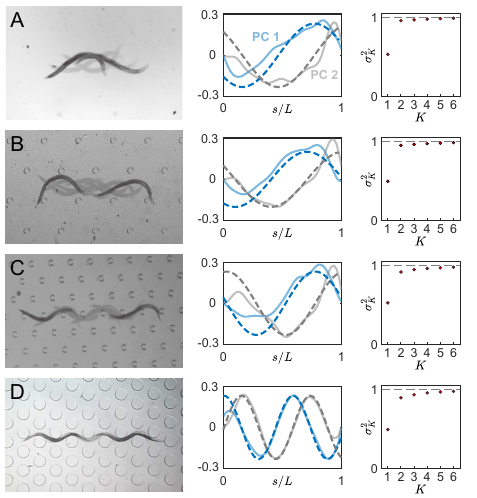}
\caption{Overlaid photos of \textit{C. elegans} movements, their low dimensional representations (principal components and shape-basis functions), and total variance explained by each principal component in (\textbf{A}) open fluid, (\textbf{B}) a sparse lattice, (\textbf{C}) a medium lattice, and (\textbf{D}) a dense lattice. In the second column, solid lines are the first two dominant PCA modes of the body curvature profile and dashed lines are their best fits to $\sin$ and $\cos$ functions. Plots in the third column show the total variance explained as a function of the number of PCs.}
\label{fig:PCs}
\end{figure}

Micro-fluidic pillar arrays were constructed using conventional soft-lithography techniques (Fig.~\ref{fig:Terrains}A). SU-8 molds were patterned via UV photolithography. Polydimethylsiloaxane or PDMS (Dow Corning Sylgard 184) was poured onto the molds (10:1 elastomer to curing agent ratio), cured at $70^{\circ}$C overnight in an oven and peeled from the molds. The PDMS devices were cut into shapes and holes for nematode loading and fluid flow were punched using a biopsy punch. Devices were then bonded to glass substrates using a handheld corona plasma treatment wand.

Microfluidic devices were first degassed using by flowing in a Pluronic and DI water mixture. Once all air was removed, the devices were flushed with flowing S-basal buffer for several minutes. Nematodes were then loaded rinsed off of their plates with S-basal, washed 3 times and loaded into a syringe. The syringe was then connected to the device and nematodes were pushed into the pillar array. The device was then sealed using capped syringe tips in the entry and exit ports and then continuously imaged for $\sim$10 minutes at 20 FPS on a dissecting scope (Leica).

\subsubsection*{Video processing}
Video recordings were first cropped to isolate bouts of individual nematodes performing bouts of forward swimming/crawling behaviors (stationary nematodes were ignored). A reference image containing only the pillars was constructed by averaging the frames of an entire bout, or by selecting a frame when the nematode was out of the cropped video. Background subtraction was then performed to isolate the nematode. Thresholding was used to binarize the image of the nematode, creating a series of black and white masks. Each mask was then skeletonized to isolate the centerline. These image processing steps were performed in ImageJ. The centerlines were then converted into curvature heatmaps in MATLAB, using a B-spline to interpolate between the pixel-wise centerline points. The curvatures were then used to perform subsequent analysis using MATLAB. 

\subsection*{Robophysical model design and manufacturing}
\begin{figure}[H]
\centering
\includegraphics[width=0.85\textwidth]{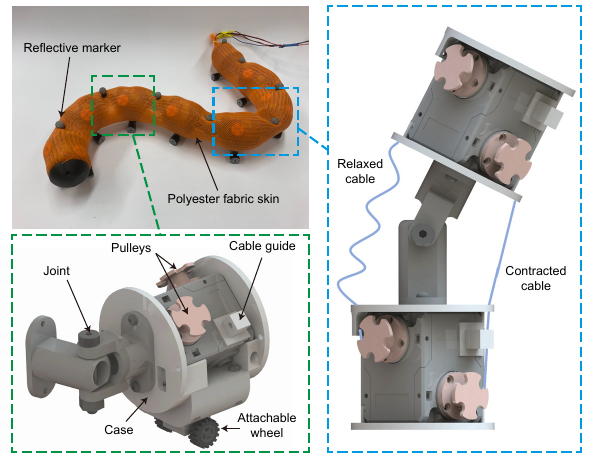}
\caption{A photo and computer aided design drawings detailing components of the robophysical model.}
\label{fig:RobotDesign}
\end{figure}

The robophysical model was constructed as a chain of linked identical modules (Fig.~\ref{fig:RobotDesign}, 7 joints and 86 cm body length). Each individual module consisted of a two-axis servo motor housed inside a case. The cases were attached to one another with a unilaterally bending joint linkage. Pulleys were then attached to each axis of the motor, and the pulleys were spooled with strings, which were referred to as cables. To complete the design, the cables were unspooled through the case and fixed onto the case ahead of the current one. Additional add-on features, such as skins and wheels, were also included for specific robophysical experiments to model the biological model.

Each module contained a Dynamixel 2XL430-W250-T servo motor (ROBOTIS), which had two axes that could be controlled independently. This feature enables the left and right cables to be adjusted to different lengths as needed. With a stall torque of 1.4 Nm, the motor provides ample support for the cable tension resulting from body-environment interactions. Additionally, the motor offers precise and continuous position control, with small enough resolution for multiple rotations. This feature allows for accurate cable length controls, where it is assumed that the cable length was approximately proportional to the motor position within the range between the maximum and minimum cable lengths.

The case that houses the servo motor serves as the main structural component and skeleton of the body. It was custom designed (55 mm length, 68 mm diamater) and manufactured to fit the motor's geometry and was 3D printed (Raise3D E2 3D printer) using PLA material. To attach the case to other components, such as the joint and wheels, heat-insets were inserted into all the holes. All the cases were identical, except for the one at the anterior end (head) of the robophysical model, which had a rounded shape for smoother head-obstacle interactions.

The joint (28 mm length) connecting adjacent modules in our system provides one degree of freedom rotation, with the axis of rotation perpendicular to the ground surface. We 3D printed joints with PLA material. Each joint allows a full range of 180 degrees of rotation, from $-90$ to $+90$ degrees, with the neutral position at $0$ degrees where the two links align. The joints are secured to the cases with two screws that connect directly to the heat insets, facilitating easy rearrangement and replacement.

The cables are the component that drives the movement of the robophysical model. To achieve this, we utilized nonelastic fishing lines (Rikimura) that boast high tensile strength of up to $180$ pounds and demonstrate negligible deformation and shape memory upon stretching. To control the shortening and lengthening of the cables, we employed pulleys (9.5-mm diameter) that were 3D printed using PLA material and attached to each rotational shaft of the servo motor. One end of each cable was fixed to the pulley, whereas the rest was tightly wound around it. This configuration allows the length of the cable to vary proportionally with the rotation angle of the pulley, which can be accurately controlled by the servo motor. The other end of each cable was threaded through a small guiding hole on the edge of the case and attached to the other case linked by the joint. For each joint, two cables were present on either side, controlling the full range of motion of the joint. A cable shortens when it is taut and under tension, whereas it lengthens when it is slack and has no tension.

Our robophysical model was controlled using code developed with the Dynamixel SDK library and programmed in MATLAB. Control signals were transmitted to the robophysical model from a PC via U2D2 (ROBOTIS). We powered the motors using a DC power supply (HY3050EX) with a voltage setting of 11.1 V. As the servo motors were connected in a daisy chain configuration for both power and communication, we connected the U2D2 and power supply to the last motor in the series.

We used an elastic mesh sleeve (1.75-inch ID polyester fabric expandable sleeving, McMaster-Carr) to wrap around the robophysical model body. Note that the sleeve cannot create anisotropy to provide any extra propulsion. The benefit of using an isotropic sleeve is twofold. The robophysical model is made of discretized hard modules and joints; therefore, it can get wedged unexpectedly in the heterogeneities because of the irregular structures, such as edges of the case. The sleeve can smooth the discretization of the body to allow for more continuous body contact with the environment. The sleeve also provides weak passive elasticity, facilitating a weak but inherent ``potential" for the robophysical model to return to the straight posture. This elasticity was found helpful especially in the passive behaviors that the robophysical model displayed and share similarities with those in biological model. The force effect of the sleeve was also considered when the force-deformation properties of the robophysical model were characterized.

The wheels are attachable components that can be attached or removed from the bottom of each case. To attach wheels onto the case, a base was 3D printed using PLA and screwed to the base. Then, the wheel frame (LEGO) was screwed into the base. The wheels were passive, non-actuated. To achieve a similar drag anisotropy for the robophysical model as for the biological model ($\sim$$1.5:1$), we replaced the rubber tires with low-friction fiberglass tape (McMaster-Carr), resulting in a $1.6:1$ drag anisotropy ($F_\perp/F_\parallel = 1.6/1$, verified with wheel force experiments following the protocols proposed in~\cite{rieser2019dynamics}). This allowed us to better model the low Reynolds number viscous fluid locomotion of the biological model. Noted that in open and sparse environments, wheels are necessary for the robophysical model to produce propulsion with drag anisotropy. However, as heterogeneity density increases, the propulsion forces provided by pushing off heterogeneity generally dominates the locomotion, and the robophysical model can move forward effectively without wheels. For consistency in the experimental setup and comparison with the biological model, we kept the wheels on for robophysical experiments in all environments.

The head collision sensor is an add-on structure in the closed-loop robophysical model, for studying how mechanical intelligence can be imposed by active reversal behaviors and modeling the head sensing neurons of \textit{C. elegans}, we designed and 3D printed a head for the robophysical model that is capable of sensing the collision angle (discrete) and the rough magnitude of collision forces. Five force-sensing resistors (FSR, Interlink Electronics FSR Model 408) were attached in parallel on the curved head surface (Fig. 7C). The feedback analog signals were collected using an Arduino micro-controller (Seeeduino XIAO SAMD21). The collision angle ranges that each FSR can detect are roughly $65^\circ$ to $75^\circ$, $75^\circ$ to $85^\circ$, $85^\circ$ to $95^\circ$, $95^\circ$ to $105^\circ$ and $105^\circ$ to $115^\circ$. The thresholds that we set to trigger the reversal behavior in the closed-loop control of the robophysical model were 3 N for the third (the middle) FSR and 5 N for the second and forth (left and right middle) FSR. When the head collision sensor sensed collision force beyond the set thresholds, the robophysical model was programmed to initiate a reversal behavior, where we fixed the reverse duration to be 0.125 cycle so that we focus on studying the effect of reversals, despite that the duration of nematode reversals was observed to vary from 0.1 to 2 cycles. 

\subsection*{Robophysical model control}
\begin{figure}[H]
\centering
\includegraphics[width=0.6\textwidth]{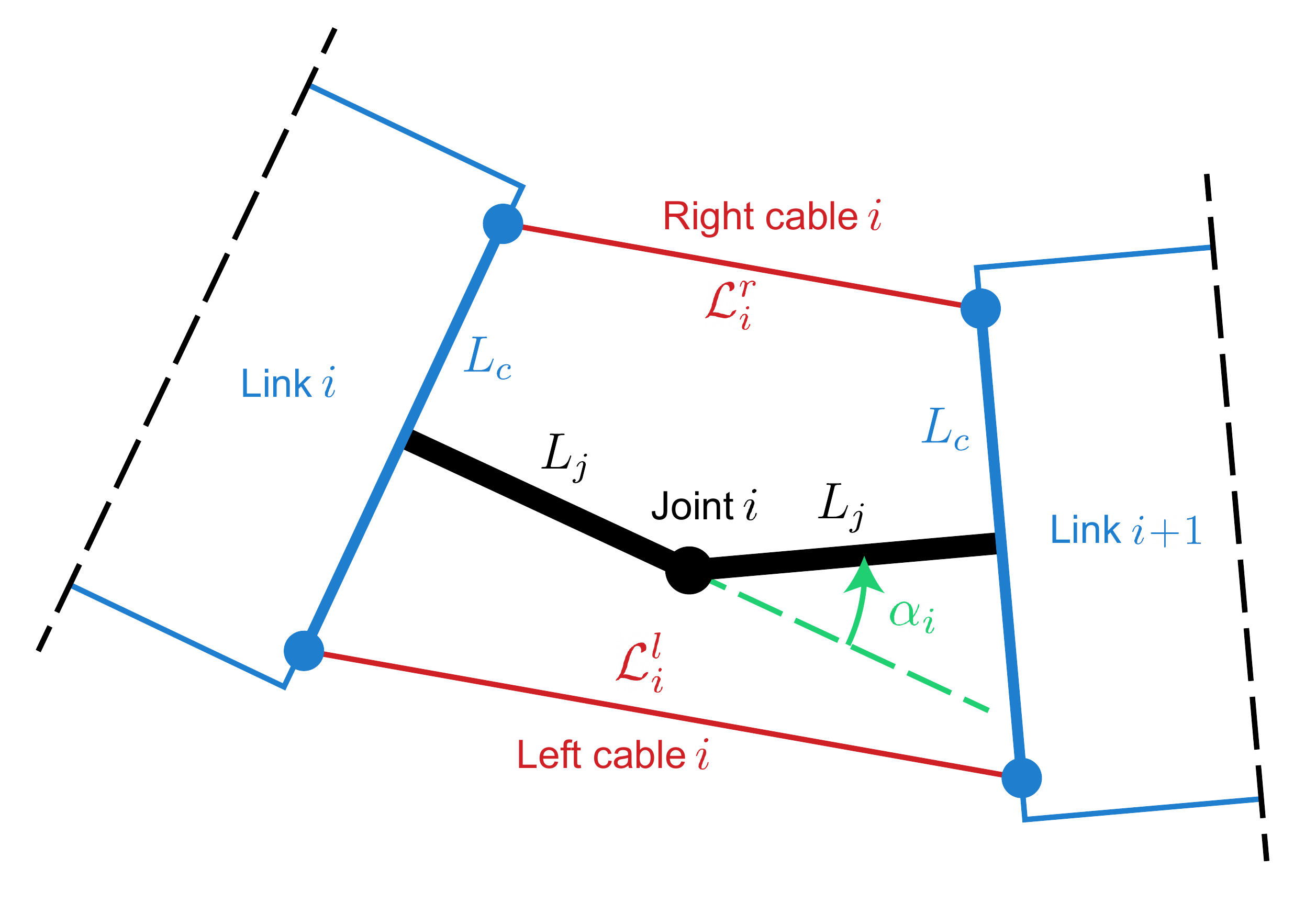}
\caption{Geometry of the joint mechanical design for the calculation of exact lengths of cables $\mathcal{L}_i^l$ and $\mathcal{L}_i^r$ to strictly form a suggested angle $\alpha_i$.}
\label{fig:JointGeometry}
\end{figure}

We calculated the exact lengths of the left and right cables that can form a joint angle $\alpha$, $\mathcal{L}^l(\alpha_i)$ and $\mathcal{L}^r(\alpha_i)$, based on the geometry of the joint mechanical design (Fig.~\ref{fig:JointGeometry}). ``Exact length" means the cable is in a shortened state, forming a straight line. Thus, $\mathcal{L}^l$ and $\mathcal{L}^r$ follow
\begin{equation}
\begin{aligned}
    \mathcal{L}^l(\alpha_i) &= 2\sqrt{L_{c}^2 + L_{j}^2} \cos\left[-\frac{\alpha_i}{2}+\tan^{-1}\left(\frac{L_{c}}{L_{j}}\right)\right],\\
    \mathcal{L}^r(\alpha_i) &= 2\sqrt{L_{c}^2 + L_{j}^2} \cos\left[\frac{\alpha_i}{2}+\tan^{-1}\left(\frac{L_{c}}{L_{j}}\right)\right].
\end{aligned}
\label{eq:ExactLength1}
\end{equation}
Considering design parameters of our robophysical model, we have
\begin{equation}
\begin{aligned}
    \mathcal{L}^l(\alpha_i) &= 79.2 \cos\left(-\frac{\alpha_i}{2}+\frac{\pi}{4}\right) \text{mm},\\
    \mathcal{L}^r(\alpha_i) &= 79.2 \cos\left(\frac{\alpha_i}{2}+\frac{\pi}{4}\right) \text{mm}.
\end{aligned}
\label{eq:ExactLength2}
\end{equation}

We followed Eq. 3 to control the lengths of the left and right cables $L_i^{l/r}$ for the $i$-th joint. We converted the linear motion of shortening and lengthening cables to the rotary motion of pulleys by spooling cables onto them. Since arc length is proportional to the center rotational angle, which we can directly control via servo motor (4096 positions per full rotation, $0.088^\circ$ resolution), we commanded the motor position $P$ to achieve the shortening and lengthening of cable length $L$ using
\begin{equation}
P(L) = P_0 - \gamma L,
\label{eq:PvsL}
\end{equation}
where $P_0$ is the position of the motor when the cable length is 0 (calibrated for each cable), and $\gamma = \frac{\text{Motor positions per full rotation}}{\text{Cable coil length per full rotation}}=\frac{4096}{\pi D_\text{pulley}} = 137.2$ mm$^{-1}$. Note that $L\geq0$ and we regulated the positive direction of motor rotation corresponds to the shortening of the cable, according to our mechanical design, thus $P_0$ is the maximum motor position and $\gamma$ is positive. Also note that, we neglected the change of pulley radii due to the thickness of the cable ($<0.5$ mm). By substituting Eq. 3 into Eq.~\ref{eq:PvsL}, we obtained the control policy in terms of motor position that we directly programmed to run the robophysical model. Practically, we set $\gamma l_0$ to be a constant with a magnitude of $100$ throughout this work, yielding $l_0=0.73$ mm/degree.

By varying the value of generalized compliance $G$, the robophysical model can display different levels of body compliance and mechanical intelligence, allowing the robophysical model to implement specific kinematics (gaits from nematodes) while passively mediate and respond to environmental perturbations. Fig.~\ref{fig:ConfigPassivity} provides a detailed explanation of the behaviors that one single joint and the whole robophysical model can display when $G$ falls in different ranges. The first schematic in each row shows the state of the joint (either bidirectionally non-compliant, directionally compliant, or bidirectionally compliant) and the state of left and right cables (either shortened or lengthened) depending on which region the suggested joint angle falls into. The second plot in each row illustrates the actual lengths comparing with the exact lengths of left and right cables on either sides of the joint as a function of the suggested joint angle, where overlaps of actual and exact lengths means the cable is shortened, whereas the discrepancy between actual and exact lengths shows how much the cable is lengthened. Note that $\mathcal{L}(0)$ on the y-axis means the exact length of a cable when joint angle is 0, $\mathcal{L}_\text{max}$ and $\mathcal{L}_\text{min}$ mean the exact length of the left (right) cable when the joint angle is $90^\circ$ and $-90^\circ$ ($-90^\circ$ and $90^\circ$), respectively. The third plot in each row illustrates the feasible range of all possible emergent joint angle, showing how much a single joint angle could depart the suggested joint angle by perturbation of external forces, enabled by lengthening of cables. The last figure in each row depicts the feasible region of all possible emergent gait paths of the robophysical model, taking all joints as a whole, in the shape space spanned by $w_1$ and $w_2$. We projected the collection of upper bounds for all joints onto the $\sin$ and $\cos$ shape basis functions to acquire the outer bound of the possible gait paths. And similarly we projected lower bounds of joint angle to acquire the inner bound of the possible gait paths. The region bounded by inner and outer bounds then illustrates how much the robophysical model could depart the suggested gait path by perturbation of external forces. 

Note that although the three representative values of $G$ ($G=0,0.5,1$) are not related to the robophysical model's geometry and gait parameter selection, the fully passive value, the value over which G exceeds the robot will become fully passive, is related to the geometry and parameter selection. The accurate fully passive value can be calculated using the forth equation given in Eq. 3,
\begin{equation}
\mathcal{L}^r[A \cdot \min(1, 2G-1)]+l_0\cdot[(2G-1)A - A] = \mathcal{L}_{max},
\end{equation}
meaning that when the commanded angle is set to the maximum amplitude ($\alpha=A$), the right cable is loosened to the maximum length such that the joint can freely bend to the minimum amplitude ($-A$); thus the joint is fully passive. Note that without the loss of symmetry, using the left cable equation (the second equation in Eq. 3) will lead to the same result. Given $G>0.5$, it can be simplified as 
\begin{equation}
\mathcal{L}^r(A)+2l_0A(G-1) = \mathcal{L}_{max}.
\end{equation}
Solve for $G$, we get $G = 1 + \frac{\mathcal{L}_{max}-\mathcal{L}^r(A)}{2l_0A}$, the fully passive value as shown in Fig.~\ref{fig:ConfigPassivity}. $\mathcal{L}_{max}$ and $\mathcal{L}^r(A)$ can be directly calculated using Eq.~\ref{eq:ExactLength2}, by letting $\alpha=\pi/2$ and $\alpha=A$. Thus, in this work, substituting in the amplitude parameters we test ($A = 46^\circ, 48^\circ, 51^\circ, 72^\circ$) and $l_0=0.73$ mm/degree, the exact fully passive values are $G = 1.74, 1.73, 1.71, 1.64$, respectively. Considering in the robophysical experiments we varied $G$ value with a $0.25$ interval, $G = 1.75$ works as a general approximated fully passive value throughout the work.

\subsection*{Robophysical kinematics analysis and comparison}

We describe the kinematics of nematodes using their curvature profile (Fig.~\ref{fig:curv}), calculated from images as described before. The local curvature is defined as $\kappa(s) = \frac{1}{r(s)}$ where $s$ is the body coordinate increasing from head to tail. 

Undulatory waves in nematodes may be approximated by a serpenoid wave~\cite{hirose1993biologically} where the curvature is a traveling wave:
\begin{equation}
    \kappa(s,t) = A\sin{(\omega t + ks)},
    \label{eq:curv_contd}
\end{equation}
where $\kappa(s,t)$ is the local curvature evaluated at time $t$ and arc-length $s$; $\omega$ is the temporal frequency and $k$ is the spatial frequency. While nematodes and other organisms are continuous, robots including our robophysical model are generally made from a small number of discrete components. To understand how the shapes of a discrete jointed undulator map onto a continuously curving undulator, we first consider the curvature of a continuous undulator evaluated at a discrete set of points along body, in which case Eq.~\ref{eq:curv_contd} can be written as
\begin{equation}
    \kappa(i,t) = A\sin{(\omega t + k_d i)},
    \label{eq:curv_dis}
\end{equation}
where $i$ is the index of discretized points. 

We further decompose the serpenoid traveling wave into the product between temporal component and spatial component:
\begin{align}
    \kappa(i,t) &= \underbrace{A\sin{(\omega t)}}_{w_1(t)} \underbrace{\cos{(k_d i})}_{\beta_1(i)} + \underbrace{A\cos{(\omega t)}}_{w_2(t)} \underbrace{\sin{(k_d i)}}_{\beta_2(i)}   \nonumber \\
    &= w_1(t)\beta_1(i) + w_2(t)\beta_2(i),
\end{align}
where $\beta_1(i)$ and $\beta_2(i)$ are time-invariant shape-basis function to prescribe a serpenoid traveling wave.

Now we consider applying the serpenoid curve to a robophysical model with discretized joints and links. Define $\vec{T}(i)$ to be the tangent vector evaluated at $i$-th points along the curve. Note that $\vec{T}(i)$ has unit length, $|\vec{T}(i)|=1$. Let $\vec{T}(i+1)$ be the unit tangent vector evaluated at $(i+1)$-th point. The distance between two consecutive points should be $\Delta s = L/N$, where $L$ is the total length of the curve and $N$ is the total number of points. Notably, $\kappa(i)$ is defined as
\begin{equation}
    \kappa(i) = \lim_{N \to \infty} \frac{|\vec{T}(i+1) - \vec{T}(i)|}{\Delta s}.
\label{eq:curve_limit}
\end{equation}
We define $\alpha(i)$ as the joint angle between the tangent vector $\vec{T}(i+1)$ and $\vec{T}(i)$. From geometry, we have
\begin{align}
    |\vec{T}(i+1) - \vec{T}(i)| =
    |\Delta \vec{T}| = 2\sin{\left(\alpha(i)/2\right)}.
\label{eq:curve_trans}
\end{align}
Substituting into Eq.~\ref{eq:curve_limit}, we have 
\begin{equation}
    \kappa(i) = \lim_{N \to \infty} \frac{2\sin{\left(\alpha(i)/2\right)}}{L/N}.
\label{eq:curve_limit_2}
\end{equation}
Since $ \lim_{N \to \infty} 2\sin{\left(\alpha(i)/2\right)} = \alpha(i)$, we have $\kappa(i) = N\alpha(i)/L$ as $N\to \infty$. Thus, in a discretized case (in our case, a robophysical model), joint angle is a reasonable alternate variable to curvature in the continuous case to describe kinematics,
\begin{equation}
    \alpha(i,t) = w_1(t) \beta_1^\alpha (i)+ w_2(t) \beta_2^\alpha (i),
\end{equation}
as in Eq.~\ref{eq:serpenoid}. Therefore, in a general sense, joint angles of the robophysical model and the body curvatures of the nematode are comparable quantities, as well as their gait paths in the shape space (as shown in Fig. 2A and Fig. 4A). More generally, continuous curvature can be mapped onto to the discrete joint angle representation of gaits. In the limit of infinite link numbers they are fully equivalent, but for finite joint numbers they coincide with points along the continuous body and only diverge between the joints.

\section*{Supplementary Discussion}

\subsection*{Coasting numbers for biological and robophysical models}
We consider fluid-swimming nematode locomotion occurs in a sufficiently low Reynolds number environment ($\sim$0.1), which permits the valid assumption of inertialess locomotion. Notably, when a nematode ceases self-deformation, its locomotory speed decays to half of its steady-state velocity in approximately 5 ms, primarily due to viscous Stokes drag~\cite{berg1993random}. We refer this period as the ``coasting time," denoted as $\tau_\text{coast}$, and introduce the dimensionless ``coasting number"~\cite{rieser2019geometric}, $\mathcal{C}=2\tau_\text{coast}/\tau_\text{cycle}$, where $\tau_\text{cycle}$ denotes the gait period, and $\tau_\text{cycle} \approx 1$ s for nematodes. Thus, $\mathcal{C}$ for nematodes is $\sim$0.01. 

We can apply the concept of inertialess locomotion to the robophysical model. To justify this extension, we assess the ratio of inertial to frictional forces in Coulomb friction-dominated systems using: $\mathcal{C}=\frac{mv_0/\tau_\text{cycle}}{\mu m g}$, where $m$, $v_0$, $\tau_\text{cycle}$, $\mu$ and $g$ are body mass, average locomotion speed, temporal gait period, friction coefficient and gravitational acceleration constants respectively. Simplifying, we obtain $\frac{v_0/(\mu g)}{\tau_\text{cycle}}$, where the numerator can be interpreted as the time required to go from steady-state locomotion to a complete stop. In frictional fluid environments, where force is approximately rate-independent, we have $\tau_\text{coast} = \frac{1}{2}v_0/(\mu g)$. In this context, this ratio for the robophysical model is then exactly $\mathcal{C}$ for nematodes. And for the robophysical model $\mathcal{C}$ is sufficiently small (on the order of 0.001), which allows us to disregard inertial effects and compare its locomotion to that of nematodes.

\subsection*{Robot performance in diverse environments}

\subsubsection*{Evaluation metrics and methods}
In addition to the wave efficiency $\eta$ (which is the ratio of the center of mass velocity to the wave propagation velocity) that we used to describe the robot's locomotion speed, we also calculated the mechanical cost of transport $\mcot$. This dimensionless quantity, widely used in the study of legged animals and robots~\cite{gabrielli1950price,collins2005efficient,seok2013design,saranli2001rhex}, gives the work required to move a unit body weight a unit distance and allows us to analyze the robot's locomotion efficiency in a more comprehensive manner.

To calculate the mechanical cost of transport, we used the formula $\mcot = W/mgd$, where $W$ is the work done by cables, $mg$ is the robot's weight, and $d$ is the distance traveled. We estimated the tension $T$ exerted by each cable using the torque sensor embedded in the servo motor (ROBOTIS 2XL430-W250-T). During an experiment, we recorded the torque readings $\tau$ from the motor with a time interval of $\Delta t = 10$ ms. To obtain the nominal torque readings $\tau_0$, which represent the ``metabolic" torques required to enable the shaft to rotate without moving the robot, we ran a calibration experiment with the same motor running the same trajectory without tying the cable to the pulley. We then estimated the tension at each time step using the formula $T = (\tau - \tau_0)/R_\text{pulley}$, where $R_\text{pulley}$ is the radius of the pulley. To estimate the distance traveled $\Delta l$, we measured the rotation angle difference $\Delta\zeta$ of the servo motor via its internal encoder within the time interval $\Delta t$ times $R_\text{pulley}$. By summing up the products of the tension and distance for each time step, we calculated the work done by one cable during an experiment. We then summed up the work done by all cables to obtain the total work done by cables. The traveled distance $d$ was measured using tracking data by summing up the distance traveled by the robot's center of geometry during each time interval.

\subsubsection*{Flat ground}
Fig.~\ref{fig:RobotCOT}A shows the robot's wave efficiency $\eta$ and mechanical cost of transport $\mcot$ on a wood-surface flat ground, where the robot was equipped with wheels to generate a $\sim$1.6:1 drag anisotropy and move forward with retrograde wave propagation along the body. Gait parameters were fixed as $A = 46^\circ$ and $\xi=0.82$ as discussed in Materials and Methods. As the generalized compliance $G$ increases, we observed a nearly proportional decrease in $\eta$ and increase in $\mcot$. We omitted data points where $\mcot>20$ in all the plots. The robot's performance on the flat ground serves as a benchmark for comparison with other environments that we tested.

\subsubsection*{Granular media}
As demonstrated in previous work, a limbless robot can generate forward thrust in granular media with retrograde wave~\cite{maladen2011undulatory,hatton2013geometric}, thus the robot was not equipped with wheels for tests in granular media. The experiments were conducted in a pool of plastic spheres with a diameter of 5 mm, which could not enter the motor and potentially damage the robot. Gait parameters were fixed as $A=60^\circ$ and $\xi=1$. At the range of $0 \leq G \leq 1$, $\eta$ shares a similar decreasing trend as on the flat ground (Fig.~\ref{fig:RobotCOT}B). Surprisingly, we observed a more dramatic decrease in the work done by cables, yielding a decreasing $\mcot$ with a local minima at $G=0.75$. From this result we posit that, with lower body compliance, much of the active work done by the robot cannot effectively transfer into thrusting forces in such environments, and is wasted instead. By increasing the body compliance to let the robot ``flow" with the environment (react to it), we reduce energy consumption without sacrificing locomotion speed. However, when $G$ is too high, the locomotion speed drops notably, leading to an increase in $\mcot$. Such a result suggests that by leveraging the mechanical intelligence in locomotion, the robot has the potential to move efficiently within granular media. 

\subsubsection*{Channel}
Channels were set up to function as models for pipes and other environments where body shapes of the robot in lateral direction are highly constrained. Previous work has modeled and demonstrated that a limbless robot can gain thrust forces purely from its interactions with walls without the need of wheels for creating drag anisotropy~\cite{chong2023gait}. Differing from nematodes using retrograde waves to move in channels~\cite{parashar2011amplitude,yuan2014gait} where we posit their thrusts primarily result from the drag anisotropy of the fluid interactions, the robot with isotropic friction needs to use direct waves to produce forward motion, solely through forces experienced on the wall. In our experiments, the robot was not equipped with wheels and we commanded the robot with a direct wave (change ``$-$" into ``$+$" in Eq. 2) with parameters $A=60^\circ$ and $\xi=1$. Specifically, the width of the robot body while employing this gait was measured as 23 cm. To make the channel a challenging environment, we set the width of the channel as 18 cm such that the robot need to ``squeeze" its body to adapt to it, which is usually the case in applications such as pipe inspection. As a result (Fig.~\ref{fig:RobotCOT}C), our robot cannot fit into the environment until $G=1$. When $G \geq 1$, the robot generated effective forward locomotion in the channel and the local minima of $\mcot$ emerged at $G = 1.25$. This result suggests that the generalized compliance $G$ enables spontaneous shape adaptation to the channel without the need of probing channel width in advance, and reduced $\mcot$ meanwhile. Notably, this conclusion holds true even for a wheeled limbless robot employing a retrograde wave with drag anisotropy.

\subsubsection*{Lattice}
In addition to $\eta$ that has been reported in the main text for the robot in regular lattices with varied density of obstacles, we evaluated $\mcot$ for all experiments (Fig.~\ref{fig:RobotCOT}D to E). As introduced in Materials and Methods, the robot was in the same condition as in experiments on the flat ground (with wheels), and executing open-loop gaits with fixed parameters obtained from direct fitting from nematode kinematics in biological experiments, $A = 48^\circ, 51^\circ, 72^\circ$ and $\xi = 0.80, 0.58, 1.02$ in sparse, medium and dense lattices, respectively.

Firstly, the obstacles in the sparse lattice impede locomotion of the robot with low $G$, resulting in reduced $\eta$ compared to that on the flat ground. However, with an increasing $G$, the more compliant robot emerged to utilized the obstacles to generate thrust by pushing off of them, leading to an improved $\eta$, known as obstacle-aided locomotion. The local minimum of $\mcot$ emerged at $G = 0.75$, where we observed both increased locomotion speed and decreased energy consumption compared to lower $G$ values.

In the medium lattice, the robot started to become ``stuck" on obstacles, where the robot cannot traverse the lattice with the commanded gait while the body was relatively rigid ($G = 0$ and $0.25$). However, under the same open-loop control for the basic pattern of head-to-tail wave propagation, locomotion emerged when the body was more compliant, where $\eta$ and $\mcot$ also reached their maximum and minimum in the range of $0.5 \leq G \leq 1$. When the body is too compliant ($G>1$), the robot cannot generate sufficient thrust, leading to a dramatic drop in $\eta$ and increase $\mcot$. 

In our experiments, we observed a similar result in the dense lattice, where only intermediate values of $G$ led to effective and efficient locomotion. Interestingly, we also noted a slight shift in the effective range of $G$ from $0.5 \leq G \leq 1$ (medium lattice) to $0.75 \leq G \leq 1.25$ (dense lattice). We posit that, with lower $G$ values, the robot is better able to generate thrust by utilizing drag anisotropy, but may struggle with adapting to the environment. On the other hand, with higher $G$ values, the robot is more compliant to the environment, but may have reduced capabilities for generating thrust (as also demonstrated by the flat ground data). As the obstacle density increases from medium to dense lattice, the constraints on body shapes become stronger, requiring the robot to be more compliant. On the other hand, in such environments, the contact forces between the robot body and the obstacles play a more dominant role in the robot's forward motion, surpassing the contribution of drag anisotropy (as evident from the robot's ability to move in the dense lattice without wheels). Therefore, higher values of $G$ are preferred in denser lattices, which explains the slight shift in the effective range of $G$ from the medium lattice to the dense lattice.

\subsubsection*{Further discussion}
In summary, our findings indicate that in highly constrained environments where interactions between the robot body and the environment play a dominant role in locomotion, an intermediate range of generalized compliance ($0.75 \leq G \leq 1.25$) enables the robot to be compliant enough to adapt to the environment, while minimizing the work required to maintain the wave propagation pattern. This results in local minima of $\mcot$, indicating an optimal balance between compliance and wave propagation efficiency. This insight sheds light on the importance of generalized compliance in enabling effective locomotion in challenging environments such as non-movable obstacles in medium/dense lattices and channels, where the robot needs to adapt its body shape to the environment while minimizing energy expenditure.

\section*{Supplementary Figures}

\begin{figure}[H]
\centering
\includegraphics[width=0.9\textwidth]{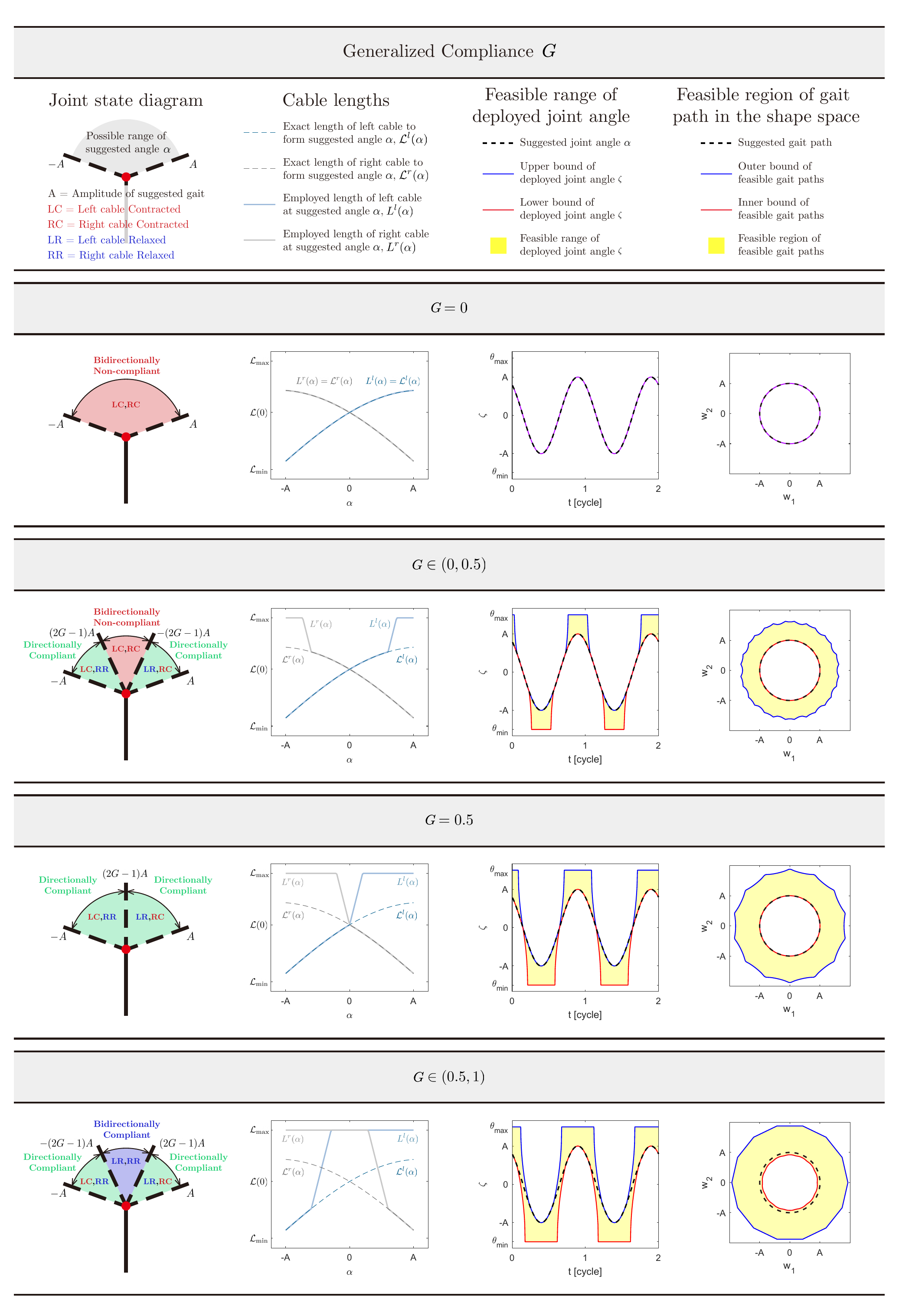}
\end{figure}

\newpage
\begin{figure}[H]
\centering
\includegraphics[width=0.9\textwidth]{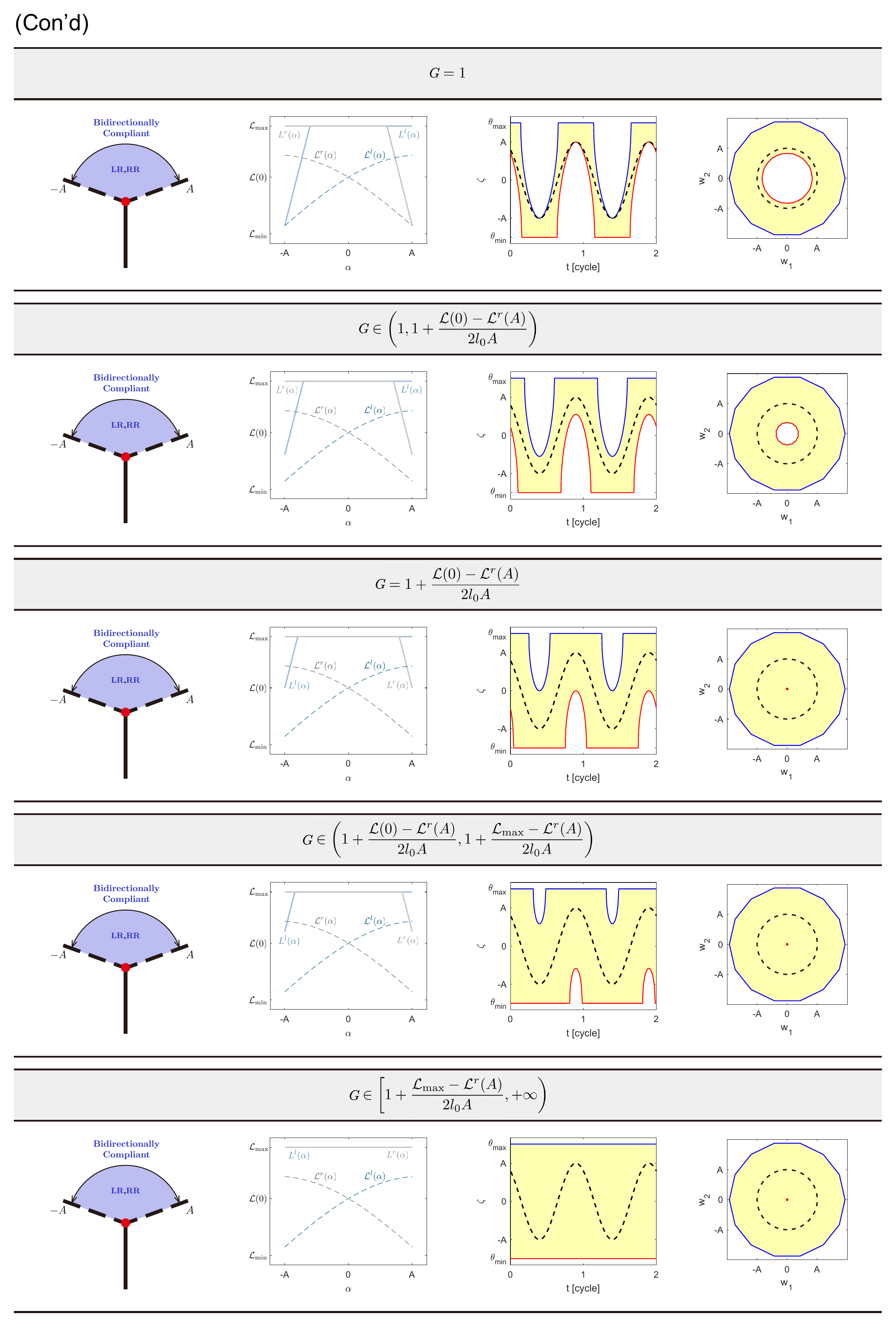}
\end{figure}

\newpage
\begin{figure} [H]
\caption{An overview of behaviors that one single joint and the whole robophysical model can display with varied generalized compliance value $G$. The first schematic in each row shows the state of the joint, left and right cables depending on which region the suggested joint angle falls into. The second plot in each row illustrates the actual lengths according to the control scheme comparing with the exact lengths of left and right cables on either sides of the joint as a function of the suggested joint angle. The third plot in each row illustrates the feasible range of all possible emergent joint angle, showing how much a single joint angle could deviate from the suggested joint angle by perturbation of external forces. The last figure in each row depicts the feasible region of all possible emergent gait paths of the robophysical model in the shape space.}
\label{fig:ConfigPassivity}
\end{figure}

\begin{figure}[H]
\centering
\includegraphics[width=0.4\textwidth]{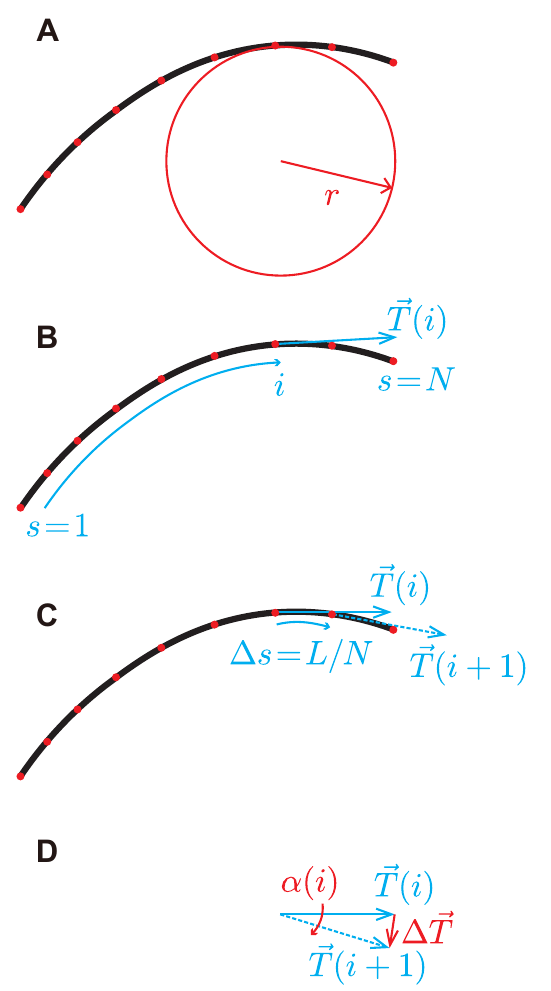}
\caption{Curvature estimation to demonstrate how discretization scheme reduces to curvature. (a) Discretization of a continuous curve and estimating the radius of curvature. (b) Tangent vector evaluated at the $i$-th point. (c) Tangent vector evaluated at the $(i+1)$-th point and the distance between two consecutive points. (d) The geometry to obtain the distance between two consecutive tangent vectors.}
\label{fig:curv}
\end{figure}

\newpage
\begin{figure}[H]
\centering
\includegraphics[width=0.6\textwidth]{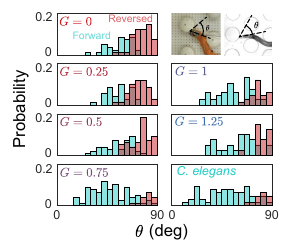}
\caption{Head collision angle probability distributions categorized by post-collision motion directions (forward or reversed) in the robophysical model with varied $G$, comparing to \textit{C. elegans} (for each plot, sample size $>100$), where the robophysical model with $G=0.75$ closely captures \textit{C. elegans} behaviors.}
\label{fig:HeadProb}
\end{figure}

\begin{figure}[H]
\centering
\includegraphics[width=0.6\textwidth]{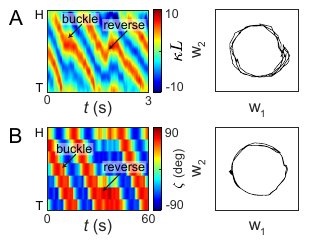}
\caption{Similar body kinematics displayed by (\textbf{A}) C. elegans and (\textbf{B}) the closed-loop robophysical model with $G=0.75$ in dense lattices, by comparing body curvature (emergent joint angles) heatmaps and gait trajectories in the shape space.}
\label{fig:RobotWormComp}
\end{figure}

\begin{figure}[H]
\centering
\includegraphics[width=0.85\textwidth]{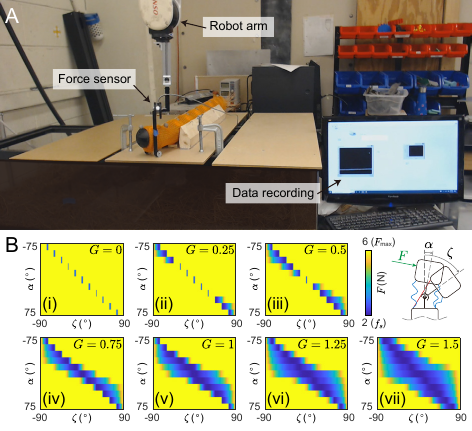}
\caption{Force-deformation property characterization for the robophysical model. (\textbf{A}) The experiment setup. (\textbf{B}) Force-deformation heatmaps for the robophysical model with varied $G$, indicating the robophysical model as a programmable functional smart material.}
\label{fig:ForceCharacterization}
\end{figure}

\begin{figure}[H]
\centering
\includegraphics[width=0.99\textwidth]{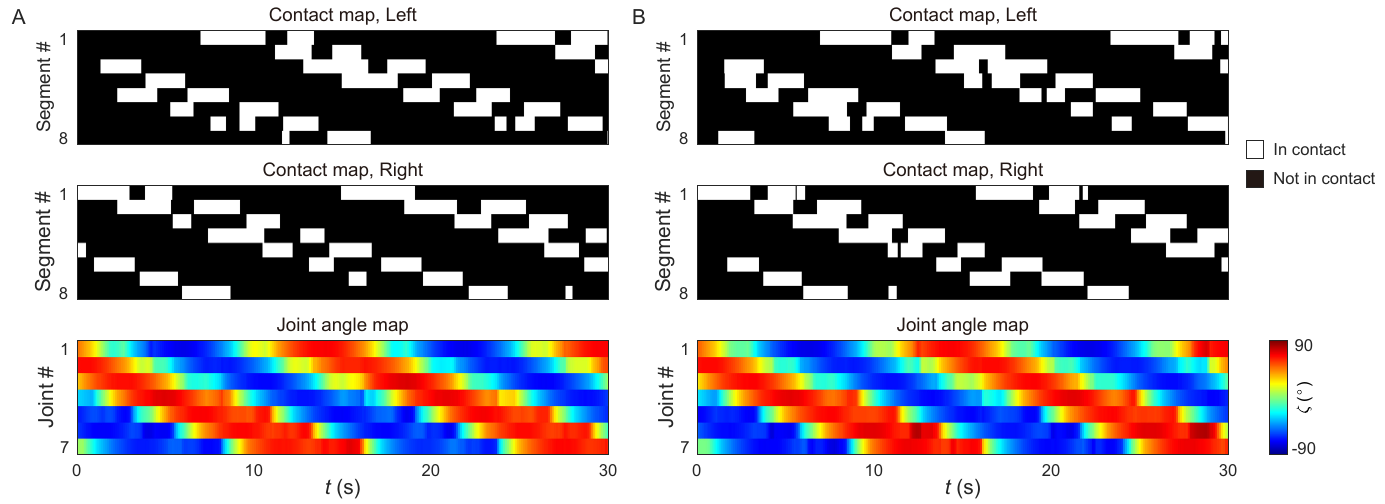}
\caption{Lattice collisions match the symmetry of the gait. Contact maps and curvature map for a wheeled (A) and wheelless (B) robot, both with $G = 0.75$. Contact maps of collisions of the robot and lattice points on the left (top row) and right (middle row) side of the body show at what body point and at what times contact with the lattice occurs (contact shown in white, absence of contact in black). These contact patterns show similar patterns to the gait, as visualized in a curvature map (bottom row) with collisions on the left-hand side of the robot corresponding with regions of positive curvature and right-hand side collisions with negative curvature. Wheeled and wheelless robots show qualitatively similar contact patterns, highlighting the dominance of lattice collisions in producing thrust (relative to ground contact) in dense lattices. Note that the head often shows longer durations of contact relative to the rest of the body, a result of the dynamics of buckling and gliding collisions.}
\label{fig:ContactMap}
\end{figure}

\begin{figure}[H]
\centering
\includegraphics[width=0.8\textwidth]{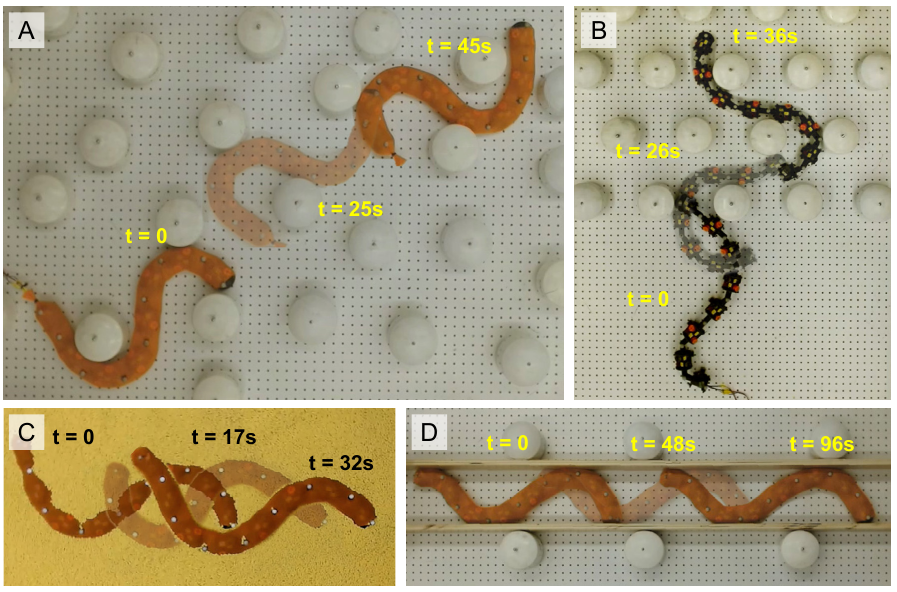}
\caption{Open-loop robotic terrestrial capabilities in various types of complex environments facilitated by mechanical intelligence. (\textbf{A}) The robot traverses a randomly distributed obstacle array. (\textbf{B}) The robot transitions from flat ground to a densely distributed obstacle array. (\textbf{C}) The robot locomotes in granular media (5 mm plastic spheres). (\textbf{D}) The robot moves in a narrow channel (18 cm width) formed with two parallel rigid walls.}
\label{fig:RobotDemos}
\end{figure}

\begin{figure}[H]
\centering
\includegraphics[width=0.95\textwidth]{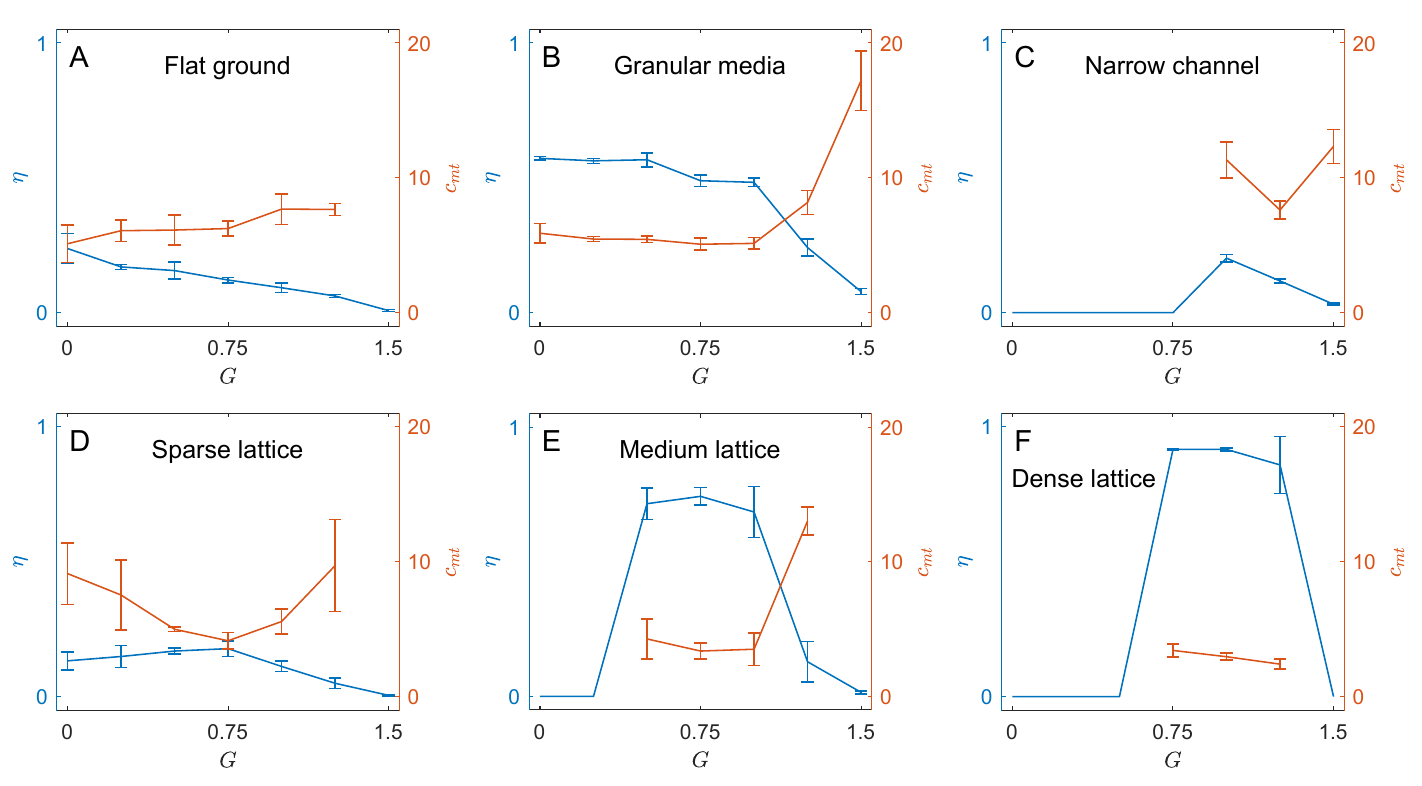}
\caption{Robot locomotion speed (wave efficiency, $\eta$) and mechanical cost of transport ($\mcot$) in different environments: (\textbf{A}) flat ground, (\textbf{B}) granular material (5 mm plastic spheres), (\textbf{C}) a narrow channel (18 cm width), (\textbf{D}) a sparse lattice, (\textbf{E}) a medium lattice, and (\textbf{F}) a dense lattice. Error bars represent standard deviations across three repetitive trials of each experiment.}
\label{fig:RobotCOT}
\end{figure}

\section*{Supplementary movie captions}

\noindent Movie S1. \textit{C. elegans} locomotion in heterogeneous terrain.

\noindent Movie S2. Overview of the robophysical model: the bilateral actuation mechanism and the programmable body compliance (generalized compliance $G$).

\noindent Movie S3. Robophysical locomotion with varied generalized compliance $G$.

\noindent Movie S4. Biological and robophysical locomotor performance comparison in all environments.

\noindent Movie S5. Biological and robophysical emergent locomotor behavior comparison.

\noindent Movie S6. Open-loop (without reversal) and closed-loop (with reversal) robophysical locomotion comparison.

\noindent Movie S7. Open-loop robotic terrestrial capabilities in complex laboratory environments, demonstrating locomotion potentials in varied environments and to broad applications.

\noindent Movie S8. Open-loop robotic terrestrial capabilities in an example outdoor complex environment, a pile of irregular rocks, demonstrating the benefit of exploiting mechanical intelligence in real-world applications.

\end{document}